\documentclass[dvipsnames]{article} 
\usepackage{iclr2023_conference,times}

\usepackage{amsmath,amsfonts,bm}

\def\eqref#1{equation~\ref{#1}}

\def\1{\bm{1}}

\def\vq{{\bm{q}}}

\def\mQ{{\bm{Q}}}
\def\mR{{\bm{R}}}

\def\mX{{\bm{X}}}

\DeclareMathAlphabet{\mathsfit}{\encodingdefault}{\sfdefault}{m}{sl}
\SetMathAlphabet{\mathsfit}{bold}{\encodingdefault}{\sfdefault}{bx}{n}

\DeclareMathOperator*{\argmax}{arg\,max}
\DeclareMathOperator*{\argmin}{arg\,min}

\usepackage{xcolor}
\usepackage[utf8]{inputenc} 
\usepackage[T1]{fontenc}    
\usepackage{hyperref}       
\usepackage{url}            
\usepackage{booktabs}       
\usepackage{amsfonts}       
\usepackage{nicefrac}       
\usepackage{microtype}      

\usepackage{amsfonts}
\usepackage{amsmath, mathtools}
\usepackage{amsthm}
\usepackage{bm}

\usepackage{algorithm}
\usepackage{subcaption}
\usepackage{algorithmicx}
\usepackage{algpseudocode}
\usepackage{wrapfig}
\allowdisplaybreaks
\usepackage{booktabs} 

\newtheorem{theorem}{Theorem}

\newtheorem{proposition}{Proposition}
\newtheorem{definition}{Definition}

\newtheorem{corollary}{Corollary}

\newcommand{\replace}[1]{{\color{Black}#1}}

 \newtheorem{innercustomprop}{Proposition}

\usepackage{tikz}
\usepackage{pgfplots}
\usetikzlibrary{shapes.geometric,arrows,arrows.meta, fit}

\newcommand{\bx}[0]{\mathbf{x}}
\newcommand{\bz}[0]{\mathbf{z}}
\newcommand{\bw}[0]{\mathbf{w}}
\newcommand{\bd}[0]{\mathbf{d}}
\newcommand{\ba}[0]{\mathbf{a}}

\newcommand{\bk}[0]{\mathbf{k}}
\newcommand{\bD}[0]{\mathbf{D}}
\newcommand{\xc}[0]{\check{\mathbf{x}}}
\newcommand{\Xc}[0]{\check{\mathbf{X}}}

\newcommand{\bdelta}[0]{\bm{\delta}}
\newcommand{\bomega}[0]{\bm{\omega}}

\newcommand{\bOmega}[0]{\mathbf{\Omega}}
\newcommand{\one}[0]{\bm{1}}
\newcommand{\bmu}[0]{\bm{\mu}}
\newcommand{\beps}[0]{\bm{\epsilon}}
\newcommand{\bX}[0]{\mathbf{X}}

\newcommand{\bI}[0]{\mathbf{I}}
\newcommand{\bH}[0]{\mathbf{H}}
\newcommand{\bK}[0]{\mathbf{K}}

\newcommand{\bW}[0]{\mathbf{W}}
\newcommand{\bY}[0]{\mathbf{Y}}
\newcommand{\bG}[0]{\mathbf{G}}

\usepackage{xspace}
\newcommand{\dataweight}{\text{data weight}\xspace}
\newcommand{\dataweights}{\text{data weights}\xspace}

\newcommand{\parameter}{\text{model parameter}\xspace}
\newcommand{\parameters}{model parameters\xspace}

\newcommand{\recourseo}{recourse outcome\xspace}
\newcommand{\recourseos}{recourse outcomes\xspace}

\newcommand{\precourse}{prescribed recourse\xspace}
\newcommand{\precourses}{prescribed recourses\xspace}

\newcommand{\ROInstability}{Recourse Outcome Instability\xspace}
\newcommand{\roinstability}{recourse outcome instability\xspace}
\newcommand{\Roinstability}{Recourse outcome instability\xspace}
\newcommand{\rainstability}{recourse action instability\xspace}
\newcommand{\Rainstability}{Recourse action instability\xspace}
\newcommand{\RAInstability}{Recourse Action Instability\xspace}

\newcommand{\delrequests}[0]{deletion requests\xspace}
\usepackage[most]{tcolorbox}

\hypersetup{
colorlinks = true,
linkcolor = ForestGreen,
anchorcolor = blue,
citecolor = Blue,
filecolor = cyan,
menucolor = ForestGreen,
runcolor = cyan,
urlcolor = ForestGreen}

\title{On the Trade-Off between  Actionable \\ Explanations and the Right to be Forgotten}

\iclrfinalcopy

\author{Martin Pawelczyk\textsuperscript{1}\thanks{Corresponding author: \href{mailto:martin.pawelczyk@uni-tuebingen.de}{martin.pawelczyk@uni-tuebingen.de}},
~ Tobias Leemann\textsuperscript{1}, 
Asia Biega\textsuperscript{2}\thanks{Equal senior author contribution.}, 
and  Gjergji Kasneci\textsuperscript{1}\footnotemark[2] \\
\textsuperscript{1}University of Tübingen, Germany \\
\textsuperscript{2}Max-Planck-Institute for Security and Privacy, Germany
\\
}

\begin{document}

\maketitle

\begin{abstract}
As machine learning (ML) models are increasingly being deployed in high-stakes applications, policymakers have suggested tighter data protection regulations (e.g., GDPR, CCPA). 
One key principle is the “right to be forgotten” which gives users the right to have their data deleted. 
Another key principle is the right to an actionable explanation, also known as algorithmic recourse, allowing users to reverse unfavorable decisions.
To date, it is unknown whether these two principles can be operationalized simultaneously.
Therefore, we introduce and study the problem of recourse invalidation in the context of data deletion requests. 
More specifically, we theoretically and empirically analyze the behavior of popular state-of-the-art algorithms and demonstrate that the recourses generated by these algorithms are likely to be invalidated if a small number of data deletion requests (e.g., 1 or 2) warrant updates of the predictive model. 
For the setting of differentiable models, we suggest a framework to identify a minimal subset of critical training points which, when removed, maximize the fraction of invalidated recourses.
Using our framework, we empirically show that the removal of as little as 2 data instances from the training set can invalidate up to 95 percent of all recourses output by popular state-of-the-art algorithms. 
Thus, our work raises fundamental questions about the compatibility of ``the right to an actionable explanation'' in the context of the ``right to be forgotten'', while also providing constructive insights on the determining factors of recourse robustness. 
\end{abstract}

\section{Introduction}
Machine learning (ML) models make a variety of consequential decisions in domains such as finance, healthcare, and policy. To protect users, laws such as the European Union's General Data Protection Regulation (GDPR)~\citep{regulation2016regulation} or the California Consumer Privacy Act (CCPA)~\citep{ccpa2021} constrain the usage of personal data and ML model deployments.
For example, individuals who have been adversely impacted by the predictions of these models have the right to \emph{recourse}~\citep{voigt2017eu}, i.e., a constructive instruction on how to act to arrive at a more desirable outcome (e.g., change a model prediction from ``loan denied'' to ``approved'').
Several approaches in recent literature tackled the problem of providing recourses by generating instance level counterfactual explanations \citep{wachter2017counterfactual,Ustun2019ActionableRI,karimi2019model,pawelczyk2019}.

Complementarily, data protection laws provide users with greater authority over their personal data.
For instance, users are granted the right to \emph{withdraw consent to the usage of their data} at any time~\citep{finckreviving}.
These regulations affect technology platforms that train their ML models on personal user data under the respective legal regime. Law scholars have argued that the continued use of ML models relying on deleted data instances could be deemed illegal \citep{villaronga2018humans}.

Irrespective of the underlying mandate, data deletion has raised a number of algorithmic research questions. In particular, recent literature has focused on the efficiency of deletion (i.e., how to delete individual data points without retraining the model \citep{ginart2019making,Golatkar_2020_CVPR}) and model accuracy aspects of data deletion (i.e., how to remove data without compromising model accuracy \citep{biega2020dm,goldsteen2021data}).
An aspect of data deletion which has not been examined before is \emph{whether and how data deletion may impact model explanation frameworks}. Thus, there is a need to understand and systematically characterize the limitations of recourse algorithms when personal user data may need to be deleted from trained ML models. Indeed, deletion of certain data instances might invalidate actionable model explanations -- both for the deleting user and, critically, unsuspecting other users. 
Such invalidations can be especially problematic in cases where users have already started to take costly actions to change their model outcomes based on previously received explanations.

In this paper, we formally examine the problem of algorithmic recourse in the context of data deletion requests.
We consider the setting where a small set of individuals has decided to withdraw their data and, as a consequence of the deletion request, the model needs to be updated \citep{ginart2019making}.
In particular, this work tackles the subsequent pressing question:

\hspace*{6mm} \emph{What is the worst impact that a deleted data instance can have on the recourse validity?}

We approach this question by considering two distinct scenarios.
The first setting considers to what extent the outdated recourses still lead to a desirable prediction (e.g., loan approval) on the updated model.
For this scenario, we suggest a robustness measure called \emph{recourse outcome instability} to quantify the fragility of recourse methods.
Second, we consider the setting where the recourse action is being updated as a consequence of the prediction model update.
In this case, we study what maximal change in recourse will be required to maintain the desirable prediction. 
To quantify the extent of this second problem, we suggest the notion of \emph{recourse action instability}.

Given these robustness measures, we derive and analyze theoretical worst-case guarantees of the maximal instability induced for linear models and neural networks in the overparameterized regime, which we study through the lens of neural tangent kernels.
We furthermore define an optimization problem for empirically quantifying recourse instability under data deletion. 
For a given trained ML model, we identify small sets of data points that maximize the proposed instability measures when deleted.
Since the resulting brute-force approach ({i.e.}, retraining models for every possible removal set) is NP-hard, we propose two relaxations for recourse instability maximization that can be optimized using (i) end-to-end gradient descent or (ii) via a greedy approximation algorithm. 
To summarize, in this work we make the following key contributions:
\begin{itemize}
\item \textbf{Novel recourse robustness problem.} We introduce the problem of \emph{recourse invalidation under the right to be forgotten} by defining two new recourse instability measures.
\item \textbf{Theoretical analysis.} Through rigorous theoretical analysis, we identify the factors that determine the instability of recourses when users whose data is part of the training set submit deletion requests.

\item \textbf{Tractable algorithms.} Using our instability measures, we present an optimization framework to identify a small set of critical training data points which, when removed, invalidates most of the issued recourses. 
\item \textbf{Comprehensive experiments.} We conduct extensive experiments on multiple real-world data sets for both regression and classification tasks with our proposed algorithms, showing that the removal of even one point from the training set can invalidate up to 95 percent of all recourses output by state-of-the-art methods
\end{itemize}
Our results also have practical implications for system designers. First, our analysis and algorithms help identify parameters and model classes leading to higher stability when a trained ML model is subjected to deletion requests.
Furthermore, our proposed methods can provide an informed way towards practical implementations of data minimization~\citep{finck2021reviving}, as one could argue that data points contributing to recourse instability could be minimized out. 
Hence, our methods could increase designer’s awareness and the compliance of their trained models.

\section{Related Work} \label{section:related_work}

\textbf{Algorithmic Approaches to Recourse.}
Several approaches in recent literature have been suggested to generate recourse for users who have been negatively impacted by model predictions \citep{tolomei2017interpretable,laugel2017inverse,Dhurandhar2018,wachter2017counterfactual,Ustun2019ActionableRI,van2019interpretable,pawelczyk2019,mahajan2019preserving,mothilal2020fat,karimi2019model,rawal2020interpretable,dandl2020multi,antoran2020getting,spooner2021counterfactual,albini2022}. 
These approaches generate recourses assuming a static environment without data deletion requests, where both the model and the recourse remain stable. 

A related line of work has focused on determining the extent to which recourses remain invariant to the model choice \citep{pawelczyk2020multiplicity,black2021consistent}, 
to data distribution shifts \citep{rawal2021modelshifts,upadhyay2021robust}, perturbations to the input instances \citep{artelt2021evaluating,dominguezolmedo2021adversarial,slack2021counterfactual}, or perturbations to the recourses \citep{pawelczyk2022noisy}.

\textbf{Sample Deletion in Predictive Models.}
Since according to EU's GDPR individuals can request to have their data deleted, several approaches in recent literature have been focusing on updating a machine learning model without the need of retraining the entire model from scratch \citep{wu2020deltagrad,ginart2019making,izzo2021,Golatkar_2020_CVPR,golatkar2020forget,cawley2004fast}.
A related line of work considers the problem of data valuation
\citep{ghorbani2020distributional, ghorbani19shapley}.
Finally, removing subsets of training data is an ingredient used for model debugging \citep{doshi2017rigorous} or the evaluation of explanation techniques \citep{hooker2019bench,rong2022evaluating}.

\textbf{Contribution.} While we do not suggest a new recourse algorithm, our work addresses the problem of recourse fragility in the presence of data deletion requests, which has previously not been studied.
To expose this fragility, we suggest effective algorithms to delete a minimal subset of critical training points so that the fraction of invalidated recourses due to a required model update is maximized.
Moreover, while prior research in the data deletion literature has primarily focused on effective data removal strategies for predictive models, there is no prior work that studies to what extent recourses output by state-of-the-art methods are affected by data deletion requests. 
Our work is the first to tackle these important problems and thereby paves the way for recourse providers to evaluate and rethink their recourse strategies in light of the right to be forgotten.

\section{Preliminaries} \label{section:preliminaries}
\vspace{-0.25cm}

\textbf{The Predictive Model and the Data Deletion Mechanism.}
We consider prediction problems from some input space $\mathbb{R}^d$ to an output space $\mathcal{Y}$, where $d$ is the number of input dimensions. 
We denote a sample by $\bz=(\bx, y)$, and denote the training data set by $\mathcal{D} = \{ \bz_1, \dots, \bz_n\}$. 
Consider the weighted empirical risk minimization problem (ERM), which gives rise to the optimal \parameters:
\begin{align}
\bw_{\bomega} = \argmin_{\bw'} \sum_{i=1}^n  \omega_i \cdot \ell\big(y_i, f_{\bw'}(\bx_i)\big),
\label{eq:weighted_erm}
\end{align}
where $\ell(\cdot, \cdot)$ is an instance-wise loss function (e.g., binary cross-entropy, mean-squared-error (MSE) loss, etc.) and $\bomega \in \{0,1\}^n$ are \dataweights that \emph{are fixed at training time}. 
If $\omega_i = 1$, then the point $\bz_i=(\bx_i, y_i)$ is part of the training data set, otherwise it is not.
During model training, we set $\omega_i = 1 ~ \forall i$, that is, the decision maker uses all available training instances at training time.
In the optimization expressed in \eqref{eq:weighted_erm}, the \parameters $\bw$ are usually an implicit function of the \dataweight vector $\bomega$ and we write $\bw_{\bomega}$ to highlight this fact; 
in particular, when all training instances are used we write $\bw_{\one}$, where $\one \in \mathbb{R}^n$ is a vector of 1s.
In summary, we have introduced the \emph{weighted} ERM problem since it allows us to understand the impact of arbitrary data deletion patterns on actionable explanations as we allow users to withdraw their entire input $\bz_i = (y_i, \bx_i)$ from the training set used to train the model $f_{\bw_{\one}}$.
Next, we present the recourse model we consider.



\textbf{The Recourse Problem in the Context of the Data Deletion Mechanism.}
We follow an established definition of counterfactual explanations originally proposed by \cite{wachter2017counterfactual}.
For a given model $f_{\bw_{\bomega}}: \mathbb{R}^d \xrightarrow{} \mathbb{R}$ parameterized by $\bw$ and a distance function $d(\cdot, \cdot): \mathcal{X} \times \mathcal{X} \to \mathbb{R}_{+}$, the problem of finding a recourse $\xc=\bx + \bdelta$ for a factual instance $\bx$ is given by:
\begin{align}
\bdelta_{\bomega,\bx} \in \argmin_{\bdelta' \in \mathcal{A}_d} \, (f_{\bw_{\bomega}}(\bx + \bdelta')-s)^{2}+\lambda \cdot \, d(\bx,\bx + \bdelta'), 
\label{eq:wachter}
\end{align}
where $\lambda \geq 0$ is a scalar tradeoff parameter and $s$ denotes the target score. 
In the optimization from \eqref{eq:wachter}, the optimal recourse action $\bdelta$ usually depends on the \parameters and since the \parameters themselves depend on the exact \dataweights configuration we write $\bdelta_{\bomega,\bx}$ to highlight this fact.
The first term in the objective on the right-hand-side of \eqref{eq:wachter} encourages the outcome $f_{\bw_{\bomega}}(\check{\mathbf{x}})$ to become close to the user-defined target score $s$, while the second term encourages the distance between the factual instance $\bx$ and the recourse $\xc_{\bomega}:= \bx + \bdelta_{\bomega, \bx}$ to be low.
The set of constraints $\mathcal{A}_d$ ensures that only admissible changes are made to the factual $\bx$. 

\textbf{Recourse Robustness Through the Lens of the Right to be Forgotten.}
We first introduce several key terms, namely, \emph{\precourses} and \emph{\recourseos}. 
A \precourse $\check{\mathbf{x}}$ refers to a recourse that was provided to an end user by a recourse method (e.g., salary was increased by \$500). 
The \recourseo $f(\check{\mathbf{x}})$ is the model's prediction evaluated at the recourse. 
With these concepts in place, we develop two recourse instability definitions.

\begin{wrapfigure}[28]{r}{0.31\textwidth}
\vspace{-0.55cm}
\centering
\begin{subfigure}{0.34\textwidth}
\centering
\begin{tikzpicture}[scale=0.70]
\draw[->] (0,0) -- (5,0) node[anchor=north] {$x_1$};

\draw node at (4, 0.5) {\small $f' = 0$};
\draw node at (+4, 4) {\small $f' = 1$};

\draw node at (4, 1.7) {\tiny \textcolor{Blue}{Original boundary: $f_{\bw_{\one}}$}};
\draw node at (4, 3.2) {\tiny \textcolor{BlueGreen}{Updated boundary: $f_{\bw_{\bomega}}$}};
\draw node at (3.8, 2.6) {\tiny \textcolor{BrickRed}{Invalidated recourse}};

\draw node at (2.3, 1) {\small $\bx$};
\coordinate (x) at (2.3, 0.85);
\fill (x) circle [radius=1pt];

\coordinate (xc) at (2.3, 2.4);
\fill (xc) circle [radius=1pt];
\draw node at (2.3, 2.6) {\small \textcolor{BrickRed}{$\tilde{\bx}_{\one}$}};

\draw[->] (2.3, 1.15) -- (2.3, 2.3);

\draw[->] (0,0) -- (0,4) node[anchor=east] {$x_2$};
\draw[thick, Blue] (0.1,0.1) -- (2,2) -- (5,2);
\draw[thick,dashed, BlueGreen] (0.1,0.1) -- (2,3) -- (5,3) ;

\vspace{-.2cm}

\end{tikzpicture}
\vspace{-0.30cm}
\caption{\ROInstability} 
\label{fig:motivation_outcome}
\end{subfigure}
\vfill 
\begin{subfigure}{0.34\textwidth}
\centering
\begin{tikzpicture}[scale=0.70]
\draw[->] (0,0) -- (5,0) node[anchor=north] {$x_1$};

\draw node at (4, 0.5) {\small $f' = 0$};
\draw node at (+4, 4) {\small $f' = 1$};

\draw node at (4, 1.7) {\tiny \textcolor{Blue}{Original boundary: $f_{\bw_{\one}}$}};
\draw node at (4, 3.2) {\tiny \textcolor{BlueGreen}{Updated boundary: $f_{\bw_{\bomega}}$}};
\draw node at (3.8, 2.6) {\tiny \textcolor{BrickRed}{Invalidated recourse}};
\draw node at (1.0, 3.2) {\tiny  \textcolor{Green}{\parbox[c]{1cm}{Updated \\ recourse}}};

\coordinate (xcn) at (1.0, 2.4);
\fill (xcn) circle [radius=1pt];
\draw node at (1.0, 2.65) {\small \textcolor{Green}{$\tilde{\bx}_{\bomega}$}};

\draw node at (2.3, 1) {\small $\bx$};
\coordinate (x) at (2.3, 0.85);
\fill (x) circle [radius=1pt];

\coordinate (xc) at (2.3, 2.4);
\fill (xc) circle [radius=1pt];
\draw node at (2.3, 2.6) {\small \textcolor{BrickRed}{$\tilde{\bx}_{\one}$}};

\draw[->] (2.3, 1.15) -- (2.3, 2.3);
\draw[->] (2.3, 1.15) -- (1.0, 2.3);

\draw[->] (0,0) -- (0,4) node[anchor=east] {$x_2$};
\draw[thick, Blue] (0.1,0.1) -- (2,2) -- (5,2);
\draw[thick,dashed, BlueGreen] (0.1,0.1) -- (2,3) -- (5,3) ;

\vspace{-.2cm}
\end{tikzpicture}
\vspace{-0.30cm}
\caption{\RAInstability} 
\label{fig:motivation_input}
\end{subfigure}
\caption{Visualizing the two key robustness notions. 
In Fig.\ \ref{fig:motivation_outcome}, recourse $\tilde{\bx}_1$ for an input $\bx$ is invalidated due to a model update. In Fig.\ \ref{fig:motivation_input}, recourse is additionally recomputed (i.e., $\tilde{\bx}_{\bomega}$) to avoid recourse invalidation.}
\label{fig:invalidation_input_regression}
\end{wrapfigure}
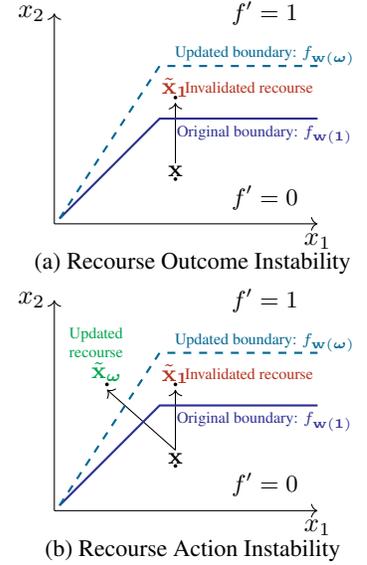

\begin{definition}(\Roinstability)
The \roinstability with respect to a factual instance $\bx$, where at least one \dataweight is set to $0$, is defined as follows:
\begin{align}
\Delta_\bx(\bomega) = \big\lvert f_{\textcolor{Blue}{\bw_{\one}}}\big(\textcolor{BrickRed}{\xc_{\one}}\big) -  f_{\textcolor{BlueGreen}{\bw_{\bomega}}}\big(\textcolor{BrickRed}{\xc_{\one}}\big) \big\rvert,
\label{eq:robustness_output}
\end{align}
where $f_{\textcolor{Blue}{\bw_{\one}}}(\textcolor{BrickRed}{\xc_{\one}})$ is the prediction at the \precourse $\textcolor{BrickRed}{\xc_{\one}}$ based on the model that uses the full training set (i.e., $f_{\textcolor{Blue}{\bw_{\one}}}$) and $f_{\textcolor{BlueGreen}{\bw_{\bomega}}}(\textcolor{BrickRed}{\xc_{\one}})$ is the prediction at the prescribed recourse for an updated model and data deletion requests have been incorporated into the predictive model (i.e., $f_{\textcolor{BlueGreen}{\bw_{\bomega}}}$).
\label{definition:ce_vulnerability_f}
\end{definition}
The above definition concisely describes the effect of applying ``outdated'' recourses to the updated model.
We assume that only the \parameters are being updated while the \precourses remain unchanged.
For a discrete model with $\mathcal{Y}= \left\{0, 1 \right\}$, Definition \ref{definition:ce_vulnerability_f} captures whether the \precourses will be invalid ($\Delta_\bx=1$) after deletion of training instances (see Fig.\ \ref{fig:motivation_outcome}). To obtain invalidation rates of recourses for a continuous-score model with target value $s$, we can also apply Definition~\ref{definition:ce_vulnerability_f} with a discretized $f'(\bx)=\mathbb{I}\left[f(\bx)>s\right]$, where $\mathbb{I}$ denotes the indicator function.

In Definition \ref{definition:ce_vulnerability_x}, consistent with related work (e.g., \cite{wachter2017counterfactual}), the distance function $d$ is specified to be a p-norm and the recourse is allowed to change due to \parameter updates.


\begin{definition}(\Rainstability)
The \Rainstability with respect to a factual input $\bx$, where at least one \dataweight is set to $0$, is defined as follows:
\begin{equation}
\Phi^{(p)}_\bx(\bomega) = \big\lVert \textcolor{BrickRed}{\xc_{\one}} - \textcolor{Green}{\xc_{\bomega}}\big\rVert_p, 
\label{eq:robustness_input}
\end{equation}
where $p \in [1, \infty)$, and $\textcolor{Green}{\xc_{\bomega}}$ is the recourse obtained for the model trained on the data instances that remain present in the data set after the deletion request.
\label{definition:ce_vulnerability_x}
\end{definition}

Definition \ref{definition:ce_vulnerability_x} quantifies the extent to which the prescribed recourses would have to additionally change to still achieve the desired \recourseo \emph{after data deletion} requests (i.e., $\xc_{\bomega}$, see Fig.\ \ref{fig:motivation_input}).
Note that we are interested in how the optimal low cost recourse changes even if the outdated recourse would remain valid.
Using our invalidation measures defined above, in the next section, we formally study the trade-offs between actionable explanations and the right to be forgotten. 
To do so, we provide data dependent upper bounds on the invalidation measures from Definitions \ref{definition:ce_vulnerability_f} and \ref{definition:ce_vulnerability_x}, which practitioners can use to probe the worst-case vulnerability of their algorithmic recourse to data deletion requests.
\section{Trade-offs Betw.\ Alg.\ Recourse and the Right to be Forgotten}
\label{section:theory}
Here we relate the two notions of recourse instability presented in Definitions \ref{definition:ce_vulnerability_f} and \ref{definition:ce_vulnerability_x} to the vulnerability of the underlying predictive model with respect to data deletion requests. We show that the introduced instability measures are directly related to data points with a high influence on the parameters after deletion.

\textbf{Analyzing the Instability Measures.} With the basic terminology in place, we provide upper bounds for the recourse instability notions defined in the previous section when the underlying models are \replace{linear} or overparameterized neural networks.
Throughout our analysis, we the term $\bd_i \coloneqq \bw_{\one} - \bw_{-i}$ has an important impact. It measures the difference from the original parameters $\bw_{\one}$ to the parameter vector $\bw_{-i}$ obtained after deleting the $i$-th instance $(\bx_i, y_i)$ from the model. In the statistics literature, this term is also known as the empirical influence function \citep{cook1980characterizations}.
Below we provide an upper bound for recourse outcome instability in linear models.

\begin{proposition}[Upper bound on recourse outcome instability \replace{for linear models}]
For the linear regression model $f(\bx) = \bw_{\text{L}}^\top \bx$ with \parameters $\bw_{\text{L}}=(\bX^{\top} \bX)^{-1} \bX^\top \bY$, an upper bound for the recourse invalidation from Definition \ref{definition:ce_vulnerability_f} by removing an instance from the training set is given by:
\begin{align}
\Delta_\bx \leq \lVert \xc_{\one} \rVert_2 \cdot \max_{i \in [n]}  ~ \lVert \bd_{i}^{\text{L}} \rVert_2,
\end{align}
where $\bd_i^{\text{L}} \coloneqq \bw_{L}{-}\bw_{L,-i} = (\bX^\top \bX)^{-1} \bx_{i} \cdot \frac{r_i}{1 - h_{ii}}$, $r_i = y_i - \bw_{\text{L}}^\top \bx_i$ and $h_{ii} = \bx_i^\top (\bX^\top \bX)^{-1} \bx_i$.
\end{proposition}
The term $(\bX^\top \bX)^{-1} \bx_{i}=\frac{d \bw_{\text{L}}}{dy_i}$ describes how sensitive the \parameters are to  $y_i$, while the residual $r_i$ captures how well $y_i$ can be fit by the model.
On the contrary, the term $h_{ii}$ from the denominator is known as the \emph{leverage} and describes how atypical $\bx_i$ is with respect to all training inputs $\bX$. 
In summary, data instances that have influential labels or are atypical
will have the highest impact when deleted.
Next, we provide a generic upper bound on recourse action instability.
\begin{proposition}[Upper bound on \rainstability]
For any predictive model with scoring function $f: \mathbb{R}^d \to \mathbb{R}$, an upper bound for the recourse instability from Definition \ref{definition:ce_vulnerability_x} by removing an instance $\bz_i=(\bx, y)$ from the training set is given by:
\begin{align}
\Phi^{(2)}_\bx &\leq \lVert \bd_i \rVert_2 \int_0^1 \bigg\lVert \frac{\bD\bdelta}{\bD\bw}\left(\tilde{\bw} \right)\bigg\rVert_2d\gamma,
\end{align}
where $\frac{\bD\bdelta}{\bD\bw}$ denotes the Jacobian of optimal recourse with the corresponding operator matrix norm, $\tilde{\bw} \coloneqq \gamma\bw +\left(1-\gamma\right)\bw_{-i}$ with $\bw_{-i}$ being the optimal model parameters with the i-th training instance removed from the training set, and $\bd_i =\bw-\bw_{-i}$.
\label{prop:upper_bound_action}
\end{proposition}
The norm of the Jacobian of optimal recourse indicates the local sensitivity of optimal recourse with respect to changes in model parameters $\bw$. High magnitudes indicate that a small change in the parameters may require a fundamentally different recourse action. The total change can be bounded by the integral over these local sensitivities, which means that low local sensitivities along the path will result in a low overall change.
Next, we specialize this result to the case of linear models.
\begin{corollary}[Upper bound on \rainstability  for linear models]
For the linear model $f(\bx) = \bw_{\text{L}}^\top \bx$ with \parameters $\bw_{\text{L}}=(\bX^\top \bX)^{-1} \bX^\top \bY$, an upper bound for the \rainstability when $s=0, \lambda \to 0$ by removing an instance from the training set is given by:
\begin{align}
\Phi^{(2)}_\bx \leq\left(\max_{i \in [n]} ~ \lVert \bd_i^{\text{L}} \rVert_2 \right)\frac{4\sqrt{2} \lVert \bx \rVert_2}{\min (\lVert \bw_{\text{L}} \rVert_2, \min_{i \in [n]}  \lVert \bw_{\text{L}, -i} \rVert_2)},
\end{align}
under the condition that $\bw_{\text{L}}^\top\bw_{\text{L}, -i} \geq 0$ (no diametrical weight changes), where $\bw_{\text{L}, -i}=\bw_{\text{L}}-\bd_i^{\text{L}}$ is the weight after removal of training instance $i$ and $\bd_i^{\text{L}}  = (\bX^T \bX)^{-1} \bx_i \frac{(y_i - \bw^\top_{\text{L}} \bx_i)}{1 - h_{ii}}$.
\end{corollary}
For models trained on large data sets, the absolute value of the \parameters' norm $\lVert \bw_L \rVert$ will not change much under deletion of a single instance. Therefore we argue that the denominator $\min (\lVert \bw_{\text{L}} \rVert_2, \min_{i \in [n]}  \lVert \bw_{\text{L}, -i} \rVert_2) \approx \lVert \bw_{\text{L}} \rVert$. Thus, \rainstability is mainly determined by the sensitivity of model parameters to deletion, $\max_{i \in [n]} ~ \lVert \bd_i^{\text{L}} \rVert_2$, scaled by the ratio of $\frac{\lVert \bx \rVert_2}{\lVert \bw_{\text{L}} \rVert_2}$.

\textbf{Neural Tangent Kernels.}
Studying the relation between deletion requests and the robustness of algorithmic recourse for models as complex as neural networks requires recent results from computational learning theory. In particular, we also rely on insights on the behaviour of over-parameterized neural networks from the theory of Neural Tangent Kernels (NTKs), which we will now briefly introduce.
Thus we study our robustness notions for neural network models in the overparameterized regime with ReLU activation functions that take the following form:
\begin{align}
f_{\text{ANN}}(\bx) = \frac{1}{\sqrt{k}} \sum_{j=1}^k a_j \cdot \text{relu}(\bw_j^\top \bx),
\end{align}
where $\bW = [\bw_1, \dots, \bw_k] \in \mathbb{R}^{d \times k}$ and $\ba = [a_1, \dots, a_k] \in \mathbb{R}^k$.
To concretely study the impact of data deletion on recourses in non-linear models such as neural networks, we leverage ideas from the neural tangent kernel (NTK) literature \citep{jacot2018neural,lee2019wide,arora2019exact,du2018gradient}.
The key insight from this literature for the purpose of our work is that infinitely wide neural networks can be expressed as a kernel ridge regression problem with the NTK under appropriate parameter initialization, and gradient descent training dynamics. 
In particular, in the limit as the number of hidden nodes $k \to \infty$, the neural tangent kernel associated with a two-layer ReLU network has a closed-form expression \citep{chen2021deep,zhang2021rethinking} (see Appendix~\ref{sec:ntk}):
\begin{align}
K^\infty(\bx_0, \bx) = \frac{\bx_0^\top \bx\left(\pi - \text{arcos}\left(\frac{\bx_0^\top \bx}{\lVert\bx_0\rVert\lVert\bx\rVert}\right)\right)}{2 \pi}.
\end{align}
Thus, the network's prediction at an input $\bx$ can be described by:
\begin{align}
f_{\text{NTK}}(\bx) & = \big(\bK^\infty(\bx, \bX)  \big)^\top \bw_{\text{NTK}},
\end{align}
where $\bX \in \mathbb{R}^{n \times d}$ is the input data matrix, $\bK^\infty(\bX, \bX)  \in \mathbb{R}^{n \times n}$ is the NTK matrix evaluated on the training data points: $[\bK^\infty(\bX, \bX) ]_{ij}=K^\infty(\bx_i, \bx_j)$ and $\bw_{\text{NTK}} = \big(\bK^\infty(\bX, \bX)  + \beta \bI_n \big)^{-1} \bY$ solves the $\ell_2$ regularized 
minimization problem with MSE loss where $\bY \in \mathbb{R}^n$ are the prediction targets. With this appropriate terminology in place we provide an upper bound on recourse outcome instability of wide neural network models.

\begin{proposition}[Upper bound on recourse outcome instability for wide neural networks]
For the NTK model with $\bw_{\text{NTK}} = \big(\bK^\infty(\bX, \bX)  + \beta \bI_n \big)^{-1} \bY$, an upper bound for the recourse invalidation from Definition \ref{definition:ce_vulnerability_f} by removing an instance $(\bx, y)$ from the training set is given by:
\begin{align}
\Delta_\bx \leq \lVert \bK^\infty(\xc_{\one}, \bX) \rVert_2 \cdot \max_{i \in [n]}  ~ \lVert \bd_{i}^{\text{NTK}} \rVert_2,
\end{align}
where $\bd_i^{\text{NTK}} = \frac{1}{k_{ii}} \bk_i \bk_i^\top \bY$, where $\bk_i$ is the $i$-th column of the matrix $\big(\bK^\infty(\bX, \bX)  + \beta \bI_n \big)^{-1}$, and $k_{ii}$ is its $i$-th diagonal element.
\end{proposition}

Intuitively, $\bd_i^{\text{NTK}}$ is the linear model analog to $\bd_i^{\text{L}}$ and $\bd_i^{\text{NTK}}$ represents the importance that the point $\bz_i = (\bx_i, y_i)$ has on the \parameters $\bw_{\text{NTK}}$.


In practical use-cases, when trying to comply with both the right to data deletion and the right to actionable explanations, our results have practical implications. For example, instances with high influence captured by $\bd_i$ should be encountered with caution during model training in order to provide reliable recourses to the individuals who seek recourse.
In summary, our results suggest that the right to data deletion may be fundamentally at odds with reliable state-of-the-art actionable explanations as the removal of an influential data instance can induce a large change in the recourse robustness, the extent to which is primarily measured by the empirical influence function $\bd_i$.


\section{Finding the Set of Most Critical Data Points} \label{section:methods}
\label{sec:optimizationalgorithms}
\textbf{The Objective Function.} In this section, we present optimization procedures that can be readily used to assess recourses'  vulnerability to deletion requests. 
On this way, we start by formulating our optimization objective.
We denote by $m \in \{\Delta, \Phi^{(2)}\}$ the measure we want to optimize for. We consider the summed instability of over the data set by omitting the subscript $\bx$, e.g., $\Delta = \sum_{\bx\in \mathcal{D}_{test}}\Delta_\bx$.
Our goal is to find the smallest number of deletion requests that leads to a maximum impact on the instability measure $m$. 
To formalize this, define the set of \dataweight configurations:
\begin{align}
\begin{split}
\Gamma_{\alpha}:= \{\bomega: \text{Maximally } \lfloor \alpha \cdot n \rfloor \text{ entries of } \bomega ~\text{are 0 and the remainder is 1.}\}.
\end{split}
\label{eq:set_data_weights}
\end{align}
In \eqref{eq:set_data_weights}, the parameter $\alpha$ controls the fraction of instances that are being removed from the training set.
For a fixed fraction $\alpha$, our problem of interest becomes:
\begin{align}
\bomega^* = \argmax_{\bomega \in \Gamma_{\alpha}}  m(\bomega).
\label{eq:optimize_invalidation} 
\end{align}
\textbf{Fundamental Problems.} When optimizing the above objective we face two fundamental problems: 
(i) \emph{evaluating} $m(\bomega)$ for many weight configurations $\bomega$ can be prohibitively expensive as the objective is defined implicitly through solutions of several non-linear optimization problems (i.e., model fitting and finding recourses). 
Further, (ii) even for an objective $m(\bomega)$ which can be computed in constant or polynomial time \emph{optimizing} this objective can still be NP-hard (a proof is given in Appendix \ref{app:np_hardness}).

\textbf{Practical Algorithms.}
We devise two practical algorithms which approach the problem in \eqref{eq:optimize_invalidation} in different ways.
As for the problem of computing $m(\bomega)$ in (i), we can either solve this by (a) using a closed-form expression indicating the dependency of $m$ on $\bomega$ or (b) by using an approximation of $m$ that is differentiable with respect to $\bomega$. 
As for the optimization in (ii), once we have established the dependency of $m$ on $\bomega$ we can either (a) use a gradient descent approach or (b) we use a greedy method.
Below we explain the individual steps required for our algorithms.
\subsection{Computing the Objective} In the objective $m(\bomega)$, notice the dependencies $\Delta_\bx(\bomega) = \Delta_{\bx}\left(f(\bw(\bomega),\check{\bx})\right)$ for the \roinstability, and  $\Phi^{(2)}_{\bx}(\bomega) = \Phi_{\bx}^{(2)}\left(\bdelta(\bw(\bomega), \bx))  \right)$ for the \rainstability. 
In the following, we briefly discuss how we efficiently compute each of these functions without numerical optimization.

\textbf{Model parameters from data weights $\bw(\bomega).$}
For the linear model, an analytical solution can be obtained,  $\bw_{\text{L}}(\bomega) = \big(\bX^\top \bOmega \bX\big)^{-1} \bX^\top \bOmega \bY$, where $\bOmega = \text{diag}(\bomega)$. The same goes for the NTK model where $\bw_{\text{NTK}}(\bomega) =  \bOmega^{\frac{1}{2}} \big(\bOmega^{\frac{1}{2}}\bK^\infty(\bX, \bX)  \bOmega^{\frac{1}{2}}  + \beta \bI\big)^{-1} \bOmega^{\frac{1}{2}} \bY$ \cite[Eqn. 3]{busuttil2007weighted}. When no closed-form expressions for the \parameters exist, we can resort to the infinitesimal jackknife (IJ) \citep{jaeckel1972infinitesimal,efron1982jackknife,giordano2019swiss,giordano2019highswiss}, that can be seen as a linear approximation to this implicit function.
We refer to Appendix \ref{appendix:implementaion_details} for additional details on this matter.

\textbf{Model prediction from model parameters $f(\bw,\check{\bx}).$} Having established the model parameters, evaluating the prediction at a given point can be quickly done even in a differentiable manner with respect to $\bw$ for the models we consider in this work.

\textbf{Recourse action from model parameters $\bdelta(\bw,\check{\bx}).$} Estimating the recourse action is more challenging as it requires solving \eqref{eq:wachter}. 
However, a differentiable solution exists for linear models, where the optimal recourse action is given by $\bdelta_{\text{L}} = \frac{s - \bw_{\text{L}}(\bomega)^\top \bx}{\lambda + \lVert \bw_{\text{L}}(\bomega) \rVert_2^2} \bw_{\text{L}}(\bomega)$. When the underlying predictor is a wide neural network we can approximate the recourse expression of the corresponding NTK, $\bdelta_{\text{NTK}} \approx \frac{s - f_{\bomega,\text{NTK}}(\bx)}{\lambda + \lVert\bar{\bw}_{\text{NTK}}(\bar{\bomega}) \rVert_2^2} \bar{\bw}_{\text{NTK}}(\bomega)$, which stems from the first-order taylor expansion $f_{\bomega,\text{NTK}}(\bx + \bdelta) \approx f_{\bomega, \text{NTK}}(\bx) + \bdelta^\top \bar{\bw}_{\text{NTK}}(\bomega)$ with $\bar{\bw}_{\text{NTK}}(\bomega) = \nabla_{\bx} K(\bx, \bX) \bw_{\text{NTK}}(\bomega)$.


\subsection{Optimizing the Objective Function}
\textbf{The Greedy Algorithm.}
We consider the model on the full data set and compute the objective function $m(\bomega)$ under deletion of every instance (alone). We then select the instance that leads to the highest increase in the objective. We add this instance to the set of deleted points. Subsequently, we refit the model and compute the impact of deletion for every second instance, when deleted in combination with the first one. Again, we add the instance that results in the largest increase to the set. Iteratively repeating these steps, we identify more instances to be deleted. Computational complexity depends on the implementation of the model weight recomputation, which is required $\mathcal{O}(\alpha n^2)$ times.

\textbf{The Gradient Descent Algorithm.} Because our developed computation of $m(\bomega)$ can be made differentiable, we also propose a \emph{gradient-based optimization} framework. We consider the  relaxation of the problem in \eqref{eq:optimize_invalidation}, 
\begin{align}
\bomega^* = \argmax_{\bomega \in \{0,1\}^{n}} ~ m(\bomega) - \lVert \one - \bomega \rVert_0,
\label{eq:ell0_objective}
\end{align}
where the $\ell_0$ norm encourages to change as few \dataweights from $1$ to $0$ as possible while few removals of training instances should have maximum impact on the robustness measure.
The problem in \eqref{eq:ell0_objective} can be further relaxed to a continuous and unconstrained optimization problem.
To do so we use a recently suggested stochastic surrogate loss for the $\ell_0$ term \citep{yamada20a}. 
Using this technique, a surrogate loss for \eqref{eq:ell0_objective} can be optimized using stochastic gradient descent (SGD). 
We refer to Appendix~\ref{appendix:implementaion_details} for more details and pseudo-code of the two algorithms. 

\section{Experimental Evaluation}  \label{section:experiments}
We experimentally evaluate our framework in terms of its ability to find significant recourse invalidations using the instability measures presented in Section \ref{section:preliminaries}.

\textbf{Data Sets.}
For our experiments on regression tasks we use two real-world data sets. In addition, we provide results for two classification datasets in the Appendix~\ref{appendix:exp_results}.
First, we use law school data from the Law School Admission Council (\emph{Admission}).
The council carried out surveys across 163 law schools in the US, in which they collected information from 21,790 law students across the US \citep{wightman1998lsac}. 
The data contains information on the students' prior performances. 
The task is to predict the students' first-year law-school average grades.
Second, we use the \emph{Home Equity Line of Credit} \emph{(Heloc)} data set.
Here, the target variable is a score indicating whether individuals will repay the Heloc account within a fixed time window.
Across both tasks we consider individuals in need of recourse if their scores lie below the median score across the data set.

\begin{figure*}[tb]
\centering
\begin{subfigure}{\textwidth}
\centering
\scalebox{0.90}{
\includegraphics[width=\textwidth]{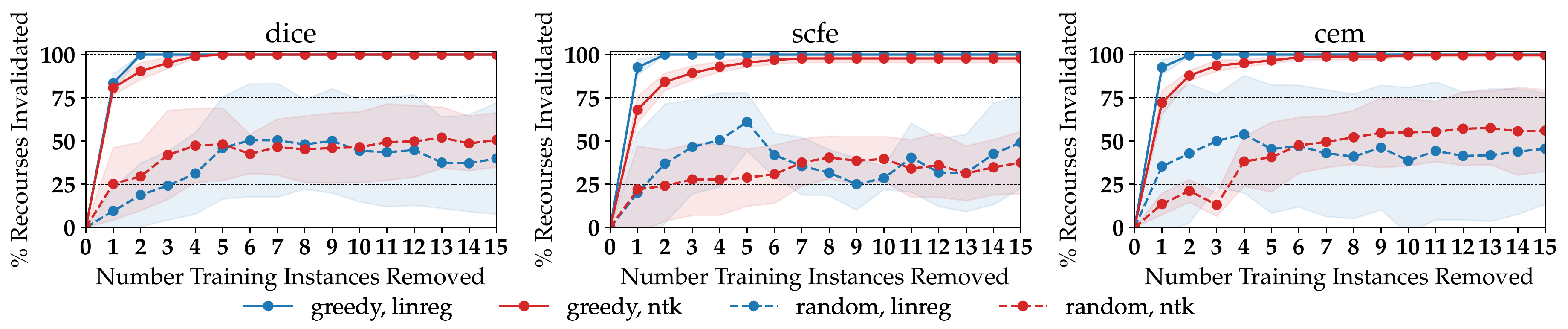}
}
\caption{Admission} 
\label{fig:invalidation_output_regression_admission}
\end{subfigure}
\vfill 
\begin{subfigure}{\textwidth}
\centering
\scalebox{0.85}{
\includegraphics[width=\textwidth]{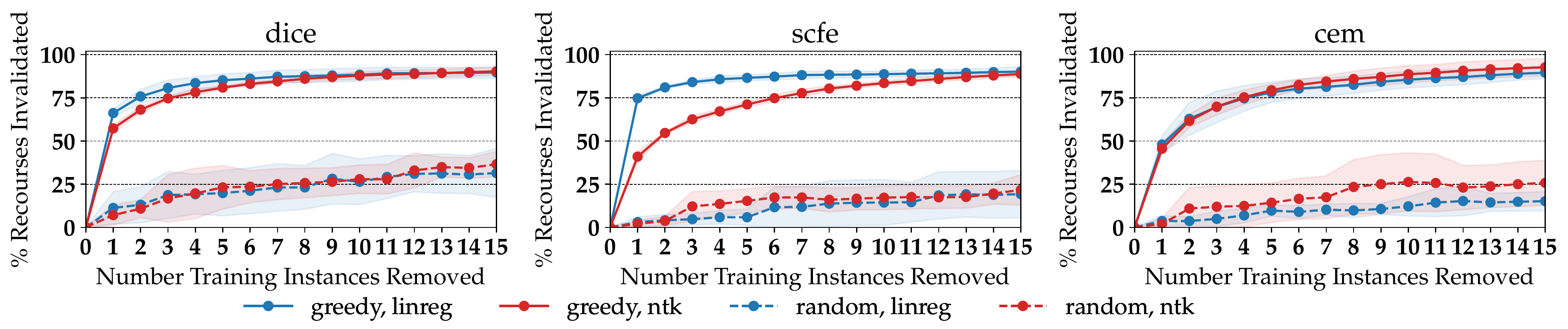}
}
\caption{Heloc} 
\label{fig:invalidation_output_regression_heloc}
\end{subfigure}
\caption{Measuring the tradeoff between \recourseo instability and the number of \delrequests for both the Admission and the Heloc data sets for regression and NTK models and various recourse methods. Results were obtained by greedy optimization; see Appendix~\ref{appendix:exp_results} for SGD results.}
\label{fig:invalidation_output_regression}
\vspace{-0.5cm}
\end{figure*}


\textbf{Recourse Methods.} We apply our techniques to four different methods which aim to generate low-cost recourses using different principles: \texttt{SCFE} was suggested by Wachter et al. \citep{wachter2017counterfactual} and uses a gradient-based objective to find recourses, \texttt{DICE} \citep{mothilal2020fat} uses a gradient-based objective to find recourses subject to a diversity constraint, and \texttt{CEM} \citep{Dhurandhar2018} uses a generative model to encourage recourses to lie on the data manifold. 
For all methods, we used the recourse method implementations from the \texttt{CARLA} library \citep{pawelczyk2021carla} and specify the $\ell_1$ cost constraint.
Further details on these algorithms are provided in App.\ \ref{appendix:implementaion_details}.


\textbf{Evaluation Measures.} For the purpose of our evaluation, we use both the \roinstability measure and the \rainstability
measure presented in Definitions \ref{definition:ce_vulnerability_f} and \ref{definition:ce_vulnerability_x}. 
We evaluate the efficacy of our framework to destabilize a large fraction of recourses using a small number of deletion requests (up to 14).
To find critical instances, we use the greedy and the gradient-based algorithms described in Sec.~\ref{sec:optimizationalgorithms}. After having established a set of critical points, we recompute the metrics with the refitted models and recourses to obtain a ground truth result.

For the \emph{\roinstability}, our metric $\Delta$ counts the number of invalidated recourses.
We use the median as the target score $s$, i.e., if the \recourseo flips back from a positive leaning prediction (above median) to a negative one (below median) it is considered invalidated. When evaluating \emph{\rainstability}, we identify a set of critical points, delete these points from the train set and refit the predictive model. In this case, we also have to recompute the recourses to evaluate $\Phi_p$.
We then measure the recourse instability using Definition \ref{definition:ce_vulnerability_x} with $p=2$.
Additionally, we compare with a random {baseline}, which deletes points uniformly at random from the train set.
We compute these measures for all individuals from the test set who require algorithmic recourse. 
To obtain standard errors, we split the test set into 5 folds and report averaged results over these 5 folds.

\textbf{Results.} In Figure \ref{fig:invalidation_output_regression}, we measure the tradeoff between \emph{\recourseo instability} and the number of deletion requests. 
We plot the number of deletion requests against the fraction of all recourses that become invalidated when up to $k \in \{1, \dots, 14\}$ training points are removed from the training set of the predictive model. 
When the underlying model is linear, we observe that the removal of as few as 5 training points induces invalidation rates of all recourses that are as high as 95 $\%$ percent -- we observe a similar trend across all recourse methods.
Note that a similar trend is present for the NTK model; however, a larger number of deletion requests (roughly 9) is required to achieve similar invalidation rates.
Finally, also note that our approach is always much more effective at deleting instances than the random baseline.
In Figure \ref{fig:invalidation_input_regression}, we measure the tradeoff between \emph{\rainstability} and the number of deletion requests with respect to the \texttt{SCFE} recourse method when the underlying predictive model is linear or an NTK model. For this complex objective, we use the more efficient SGD optimization.
Again, we observe that our optimization method significantly outperforms the random baselines at finding the most influential points to be removed.

\textbf{Additional Models and Tasks.} In addition to the here presented results, we provide results for classification tasks with (a) Logistic Regression, (b) Kernel-SVM and (c) ANN models on two additional data sets (Diabetes and COMPAS) in Appendix~\ref{appendix:exp_results}. 
Across all these models, we observe that our removal algorithms outperform random guessing; often by up to 75 percentage points.

\textbf{Factors of Recourse Robustness.} Our empirical results shed light on which factors are influential in determining robustness of trained ML models with respect to deletion requests. In combination with results from Fig.\ \ref{fig:invalidation_output_classification_greedy} (see Appendix \ref{appendix:exp_results}), our results suggest that linear models are more susceptible to invalidation in the worst-case but are slightly more robust when it comes to random removals. Furthermore, the characteristics of the data set play a key role; in particular those of the critical points.
We perform an additional experiment where we consider modified data sets without the most influential points identified by our optimization approaches. 
In Appendix \ref{appendix:exp_results}, initial results show that this simple technique decreases the invalidation probabilities by up to 6 percentage points.

\begin{figure*}[tb]
\centering
\begin{subfigure}{0.45\textwidth}
\centering
\scalebox{0.75}{
\includegraphics[width=\textwidth]{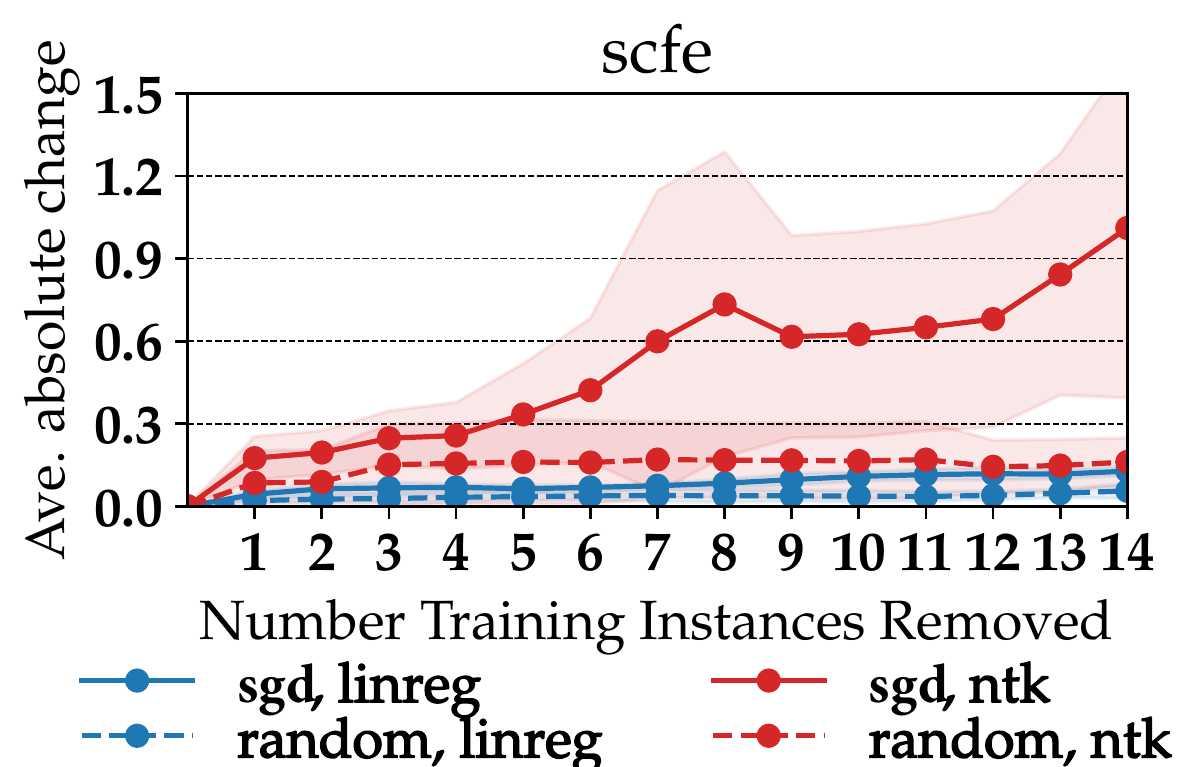}
}
\caption{Admission} 
\label{fig:invalidation_input_regression_admission}
\end{subfigure}
\hfill
\begin{subfigure}{0.45\textwidth}
\centering
\scalebox{0.75}{
\includegraphics[width=\textwidth]{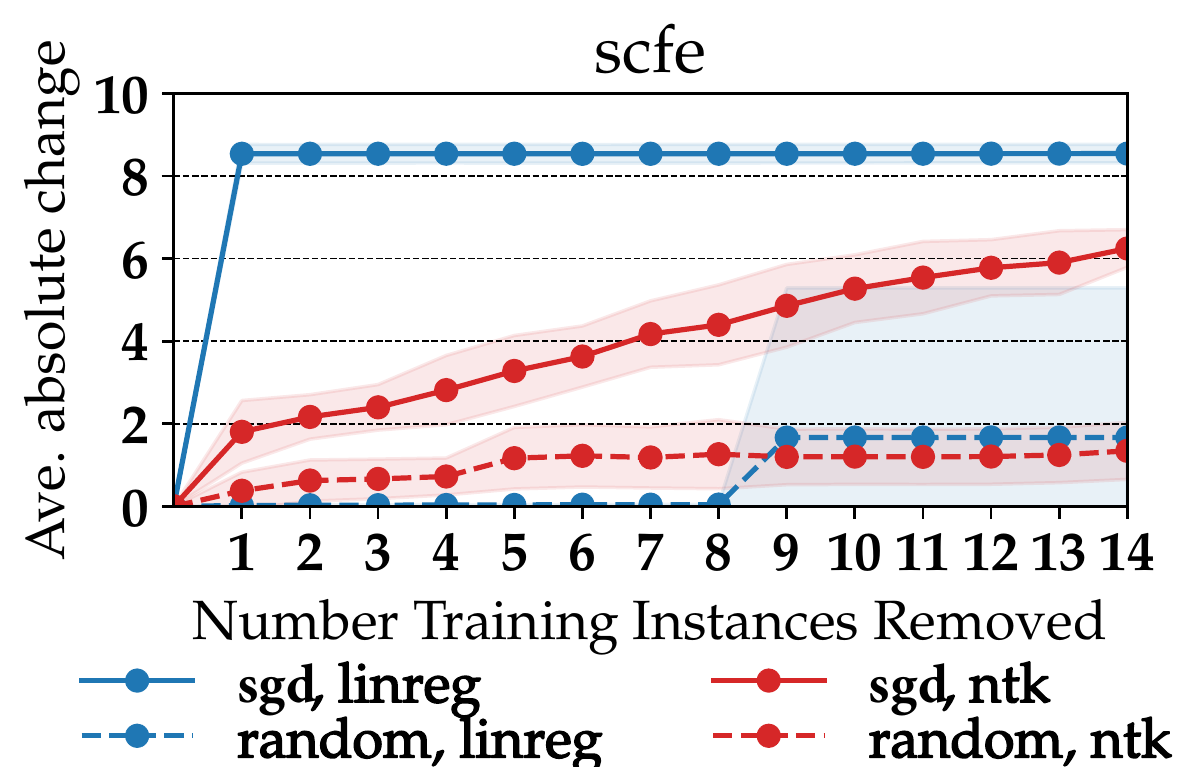}
}
\caption{Heloc} 
\label{fig:invalidation_input_regression_heloc}
\end{subfigure}
\caption{Quantifying the tradeoff between \rainstability as measured in Definition \ref{definition:ce_vulnerability_x} and the number of \delrequests for both the Admission and the Heloc data sets for the \texttt{SCFE} method when the underlying model is linear or an NTK (results by SGD optimization).}
\label{fig:invalidation_input_regression}
\vspace{-0.5cm}
\end{figure*}

\section{Discussion and Conclusion}  \label{section:conlusion}
In this work, we made the first step towards understanding the tradeoffs between actionable model explanations and the right to be forgotten.
We theoretically analyzed the robustness of state-of-the-art recourse methods under data deletion requests and suggested (i) a greedy and (ii) a gradient-based algorithm to efficiently identify a small subset of individuals, whose data, when removed, would lead to invalidation of a large number of recourses for unsuspecting other users.
Our experimental evaluation with multiple real-world data sets on both regression and classification tasks demonstrates that the right to be forgotten presents a significant challenge to the reliability of actionable explanations. 

Finally, our theoretical results suggest that the robustness to deletion increases when the model parameter changes under data deletion remain small. 
This formulation closely resembles the definition of \emph{Differential Privacy} (DP) \citep{ dwork2014algorithmic}.
We therefore conjecture that the reliability of actionable recourse could benefit from models that have been trained under DP constraints.
As the field of AI rapidly evolves, data protection authorities will further refine the precise interpretations of general principles in regulations such as GDPR. 
The present paper contributes towards this goal theoretically, algorithmically, and empirically by providing evidence of tensions between different data protection principles.

\textbf{Ethics statement.}
Our findings raise compelling questions on the deployment of counterfactual explanations in practice.
First of all, \emph{Are the two requirements of actionable explanations and the right to be forgotten fundamentally at odds with one another?} The theoretical and empirical results in this work indicate that for many model and recourse method pairs, this might indeed be the case. 
This finding leads to the pressing follow-up question: \emph{How can practitioners make sure that their recourses stay valid under deletion requests?}
A first take might be to implement the principle of data minimization \citep{biega2020dm, finckreviving,shanmugam2022dm} in the first place, i.e., exclude the $k$ most critical data points from model training.

\textbf{Acknowledgments.}
We would like to thank the anonymous reviewers for their insightful feedback.
MP would like to thank Jason Long, Emanuele Albini, Jiāháo Chén, Daniel Dervovic and Daniele Magazzeni for insightful discussions at early stages of this work.

\bibliography{bibfile.bib}

\begin{thebibliography}{67}
\providecommand{\natexlab}[1]{#1}
\providecommand{\url}[1]{\texttt{#1}}
\expandafter\ifx\csname urlstyle\endcsname\relax
  \providecommand{\doi}[1]{doi: #1}\else
  \providecommand{\doi}{doi: \begingroup \urlstyle{rm}\Url}\fi

\bibitem[Albini et~al.(2022)Albini, Long, Dervovic, and Magazzeni]{albini2022}
Emanuele Albini, Jason Long, Danial Dervovic, and Daniele Magazzeni.
\newblock Counterfactual shapley additive explanations.
\newblock In \emph{Proceedings of the Conference on Fairness, Accountability,
  and Transparency (FAccT)}, 2022.

\bibitem[Angwin et~al.(2016)Angwin, Larson, Mattu, and Kirchner]{Angwin2016}
Julia Angwin, Jeff Larson, Surya Mattu, and Lauren Kirchner.
\newblock Machine bias: There’s software used across the country to predict
  future criminals. and it’s biased against blacks, 2016.

\bibitem[Antorán et~al.(2021)Antorán, Bhatt, Adel, Weller, and
  Hernández-Lobato]{antoran2020getting}
Javier Antorán, Umang Bhatt, Tameem Adel, Adrian Weller, and José~Miguel
  Hernández-Lobato.
\newblock Getting a clue: A method for explaining uncertainty estimates.
\newblock In \emph{International Conference on Learning Representations
  (ICLR)}, 2021.

\bibitem[Arora et~al.(2019)Arora, Du, Hu, Li, Salakhutdinov, and
  Wang]{arora2019exact}
Sanjeev Arora, Simon~S Du, Wei Hu, Zhiyuan Li, Russ~R Salakhutdinov, and
  Ruosong Wang.
\newblock On exact computation with an infinitely wide neural net.
\newblock \emph{Advances in Neural Information Processing Systems (NeurIPS)},
  32, 2019.

\bibitem[Artelt et~al.(2021)Artelt, Vaquet, Velioglu, Hinder, Brinkrolf,
  Schilling, and Hammer]{artelt2021evaluating}
André Artelt, Valerie Vaquet, Riza Velioglu, Fabian Hinder, Johannes
  Brinkrolf, Malte Schilling, and Barbara Hammer.
\newblock Evaluating robustness of counterfactual explanations.
\newblock \emph{arXiv:2103.02354}, 2021.

\bibitem[Biega \& Finck(2021)Biega and Finck]{finckreviving}
Asia~J. Biega and Mich{\`e}le Finck.
\newblock Reviving purpose limitation and data minimisation in data-driven
  systems.
\newblock \emph{Technology and Regulation}, 2021.

\bibitem[Biega et~al.(2020)Biega, Potash, Daum\'{e}, Diaz, and
  Finck]{biega2020dm}
Asia~J. Biega, Peter Potash, Hal Daum\'{e}, Fernando Diaz, and Mich\`{e}le
  Finck.
\newblock Operationalizing the legal principle of data minimization for
  personalization.
\newblock In \emph{ACM(43) SIGIR ’20}, pp.\  399–408, 2020.

\bibitem[Black et~al.(2021)Black, Wang, Fredrikson, and
  Datta]{black2021consistent}
Emily Black, Zifan Wang, Matt Fredrikson, and Anupam Datta.
\newblock Consistent counterfactuals for deep models.
\newblock \emph{arXiv:2110.03109}, 2021.

\bibitem[Busuttil \& Kalnishkan(2007)Busuttil and
  Kalnishkan]{busuttil2007weighted}
Steven Busuttil and Yuri Kalnishkan.
\newblock Weighted kernel regression for predicting changing dependencies.
\newblock In \emph{European Conference on Machine Learning}, pp.\  535--542.
  Springer, 2007.

\bibitem[Cawley \& Talbot(2004)Cawley and Talbot]{cawley2004fast}
Gavin~C Cawley and Nicola~LC Talbot.
\newblock Fast exact leave-one-out cross-validation of sparse least-squares
  support vector machines.
\newblock \emph{Neural networks}, 17\penalty0 (10):\penalty0 1467--1475, 2004.

\bibitem[Chen \& Xu(2021)Chen and Xu]{chen2021deep}
Lin Chen and Sheng Xu.
\newblock Deep neural tangent kernel and laplace kernel have the same
  {\{}rkhs{\}}.
\newblock In \emph{International Conference on Learning Representations
  (ICLR)}, 2021.

\bibitem[Cho \& Saul(2009)Cho and Saul]{cho2009kernel}
Youngmin Cho and Lawrence Saul.
\newblock Kernel methods for deep learning.
\newblock \emph{Advances in neural information processing systems (NeurIPS)},
  22, 2009.

\bibitem[Cook \& Weisberg(1980)Cook and Weisberg]{cook1980characterizations}
R~Dennis Cook and Sanford Weisberg.
\newblock Characterizations of an empirical influence function for detecting
  influential cases in regression.
\newblock \emph{Technometrics}, 22\penalty0 (4):\penalty0 495--508, 1980.

\bibitem[Csat{\'o} \& Opper(2002)Csat{\'o} and Opper]{csato2002sparse}
Lehel Csat{\'o} and Manfred Opper.
\newblock Sparse on-line gaussian processes.
\newblock \emph{Neural computation}, 14\penalty0 (3):\penalty0 641--668, 2002.

\bibitem[Dandl et~al.(2020)Dandl, Molnar, Binder, and Bischl]{dandl2020multi}
Susanne Dandl, Christoph Molnar, Martin Binder, and Bernd Bischl.
\newblock Multi-objective counterfactual explanations.
\newblock In \emph{International Conference on Parallel Problem Solving from
  Nature}, pp.\  448--469. Springer, 2020.

\bibitem[Dhurandhar et~al.(2018)Dhurandhar, Chen, Luss, Tu, Ting, Shanmugam,
  and Das]{Dhurandhar2018}
Amit Dhurandhar, Pin-Yu Chen, Ronny Luss, Chun-Chen Tu, Paishun Ting,
  Karthikeyan Shanmugam, and Payel Das.
\newblock Explanations based on the missing: Towards contrastive explanations
  with pertinent negatives.
\newblock In \emph{Advances in Neural Information Processing Systems
  (NeurIPS)}, 2018.

\bibitem[Dominguez-Olmedo et~al.(2022)Dominguez-Olmedo, Karimi, and
  Schölkopf]{dominguezolmedo2021adversarial}
Ricardo Dominguez-Olmedo, Amir-Hossein Karimi, and Bernhard Schölkopf.
\newblock On the adversarial robustness of causal algorithmic recourse.
\newblock In \emph{International Conference on Machine Learning (ICML)}, 2022.

\bibitem[Doshi-Velez \& Kim(2017)Doshi-Velez and Kim]{doshi2017rigorous}
Finale Doshi-Velez and Been Kim.
\newblock Towards a rigorous science of interpretable machine learning.
\newblock 2017.

\bibitem[Du et~al.(2019)Du, Zhai, Poczos, and Singh]{du2018gradient}
Simon~S. Du, Xiyu Zhai, Barnabas Poczos, and Aarti Singh.
\newblock Gradient descent provably optimizes over-parameterized neural
  networks.
\newblock In \emph{International Conference on Learning Representations
  (ICLR)}, 2019.

\bibitem[Dwork et~al.(2014)Dwork, Roth, et~al.]{dwork2014algorithmic}
Cynthia Dwork, Aaron Roth, et~al.
\newblock The algorithmic foundations of differential privacy.
\newblock \emph{Found. Trends Theor. Comput. Sci.}, 9\penalty0 (3-4):\penalty0
  211--407, 2014.

\bibitem[Efron(1982)]{efron1982jackknife}
Bradley Efron.
\newblock \emph{The jackknife, the bootstrap and other resampling plans}.
\newblock SIAM, 1982.

\bibitem[Finck \& Biega(2021)Finck and Biega]{finck2021reviving}
Michele Finck and Asia~J Biega.
\newblock Reviving purpose limitation and data minimisation in data-driven
  systems.
\newblock \emph{Technology and Regulation}, 2021:\penalty0 44--61, 2021.

\bibitem[Ghorbani \& Zou(2019)Ghorbani and Zou]{ghorbani19shapley}
Amirata Ghorbani and James Zou.
\newblock Data shapley: Equitable valuation of data for machine learning.
\newblock In Kamalika Chaudhuri and Ruslan Salakhutdinov (eds.),
  \emph{Proceedings of the 36th International Conference on Machine Learning
  (ICML)}, volume~97, pp.\  2242--2251. PMLR, 2019.

\bibitem[Ghorbani et~al.(2020)Ghorbani, Kim, and
  Zou]{ghorbani2020distributional}
Amirata Ghorbani, Michael Kim, and James Zou.
\newblock A distributional framework for data valuation.
\newblock In \emph{International Conference on Machine Learning (ICML)}, pp.\
  3535--3544. PMLR, 2020.

\bibitem[Ginart et~al.(2019)Ginart, Guan, Valiant, and Zou]{ginart2019making}
Antonio Ginart, Melody~Y. Guan, Gregory Valiant, and James Zou.
\newblock Making ai forget you: Data deletion in machine learning.
\newblock In \emph{Proceedings of the 33rd Conference on Neural Information
  Processing Systems (NeurIPS 2019), Vancouver, Canada}, 2019.

\bibitem[Giordano et~al.(2019{\natexlab{a}})Giordano, Jordan, and
  Broderick]{giordano2019highswiss}
Ryan Giordano, Michael~I. Jordan, and Tamara Broderick.
\newblock A higher-order swiss army infinitesimal jackknife.
\newblock \emph{ArXiv}, abs/1907.12116, 2019{\natexlab{a}}.

\bibitem[Giordano et~al.(2019{\natexlab{b}})Giordano, Stephenson, Liu, Jordan,
  and Broderick]{giordano2019swiss}
Ryan Giordano, William Stephenson, Runjing Liu, Michael Jordan, and Tamara
  Broderick.
\newblock A swiss army infinitesimal jackknife.
\newblock In \emph{Proceedings of the Twenty-Second International Conference on
  Artificial Intelligence and Statistics (AISTATS)}, 2019{\natexlab{b}}.

\bibitem[Golatkar et~al.(2020{\natexlab{a}})Golatkar, Achille, and
  Soatto]{Golatkar_2020_CVPR}
Aditya Golatkar, Alessandro Achille, and Stefano Soatto.
\newblock Eternal sunshine of the spotless net: Selective forgetting in deep
  networks.
\newblock In \emph{Proceedings of the IEEE/CVF Conference on Computer Vision
  and Pattern Recognition (CVPR)}, June 2020{\natexlab{a}}.

\bibitem[Golatkar et~al.(2020{\natexlab{b}})Golatkar, Achille, and
  Soatto]{golatkar2020forget}
Aditya Golatkar, Alessandro Achille, and Stefano Soatto.
\newblock Forgetting outside the box: Scrubbing deep networks of information
  accessible from input-output observations.
\newblock \emph{arXiv:2003.02960}, 2020{\natexlab{b}}.

\bibitem[Goldsteen et~al.(2021)Goldsteen, Ezov, Shmelkin, Moffie, and
  Farkash]{goldsteen2021data}
Abigail Goldsteen, Gilad Ezov, Ron Shmelkin, Micha Moffie, and Ariel Farkash.
\newblock Data minimization for gdpr compliance in machine learning models.
\newblock \emph{AI and Ethics}, pp.\  1--15, 2021.

\bibitem[Hooker et~al.(2019)Hooker, Erhan, Kindermans, and
  Kim]{hooker2019bench}
Sara Hooker, Dumitru Erhan, Pieter-Jan Kindermans, and Been Kim.
\newblock A benchmark for interpretability methods in deep neural networks.
\newblock In \emph{Advances in Neural Information Processing Systems
  (NeurIPS)}, 2019.

\bibitem[Izzo et~al.(2021)Izzo, Anne~Smart, Chaudhuri, and Zou]{izzo2021}
Zachary Izzo, Mary Anne~Smart, Kamalika Chaudhuri, and James Zou.
\newblock Approximate data deletion from machine learning models.
\newblock In Arindam Banerjee and Kenji Fukumizu (eds.), \emph{Proceedings of
  The 24th International Conference on Artificial Intelligence and Statistics
  (AISTATS)}, volume 130. PMLR, 2021.

\bibitem[Jacot et~al.(2018)Jacot, Gabriel, and Hongler]{jacot2018neural}
Arthur Jacot, Franck Gabriel, and Cl{\'e}ment Hongler.
\newblock Neural tangent kernel: Convergence and generalization in neural
  networks.
\newblock \emph{Advances in neural information processing systems (NeurIPS)},
  31, 2018.

\bibitem[Jaeckel(1972)]{jaeckel1972infinitesimal}
L~Jaeckel.
\newblock The infinitesimal jackknife. memorandum.
\newblock Technical report, MM 72-1215-11, Bell Lab. Murray Hill, NJ, 1972.

\bibitem[Karimi et~al.(2020)Karimi, Barthe, Balle, and Valera]{karimi2019model}
Amir-Hossein Karimi, Gilles Barthe, Borja Balle, and Isabel Valera.
\newblock Model-agnostic counterfactual explanations for consequential
  decisions.
\newblock In \emph{International Conference on Artificial Intelligence and
  Statistics (AISTATS)}, 2020.

\bibitem[Karp(1972)]{karp1972reducibility}
Richard~M Karp.
\newblock Reducibility among combinatorial problems.
\newblock In \emph{Complexity of computer computations}, pp.\  85--103.
  Springer, 1972.

\bibitem[Laugel et~al.(2017)Laugel, Lesot, Marsala, Renard, and
  Detyniecki]{laugel2017inverse}
Thibault Laugel, Marie-Jeanne Lesot, Christophe Marsala, Xavier Renard, and
  Marcin Detyniecki.
\newblock Inverse classification for comparison-based interpretability in
  machine learning.
\newblock \emph{arXiv preprint arXiv:1712.08443}, 2017.

\bibitem[Lee et~al.(2019)Lee, Xiao, Schoenholz, Bahri, Novak, Sohl-Dickstein,
  and Pennington]{lee2019wide}
Jaehoon Lee, Lechao Xiao, Samuel Schoenholz, Yasaman Bahri, Roman Novak, Jascha
  Sohl-Dickstein, and Jeffrey Pennington.
\newblock Wide neural networks of any depth evolve as linear models under
  gradient descent.
\newblock \emph{Advances in neural information processing systems (NeurIPS)},
  32, 2019.

\bibitem[Mahajan et~al.(2019)Mahajan, Tan, and Sharma]{mahajan2019preserving}
Divyat Mahajan, Chenhao Tan, and Amit Sharma.
\newblock Preserving causal constraints in counterfactual explanations for
  machine learning classifiers.
\newblock \emph{arXiv preprint arXiv:1912.03277}, 2019.

\bibitem[Miller~Jr(1974)]{miller1974unbalanced}
Rupert~G Miller~Jr.
\newblock An unbalanced jackknife.
\newblock \emph{The Annals of Statistics}, pp.\  880--891, 1974.

\bibitem[Mothilal et~al.(2020)Mothilal, Sharma, and Tan]{mothilal2020fat}
Ramaravind~K. Mothilal, Amit Sharma, and Chenhao Tan.
\newblock Explaining machine learning classifiers through diverse
  counterfactual explanations.
\newblock In \emph{Proceedings of the Conference on Fairness, Accountability,
  and Transparency (FAT*)}, 2020.

\bibitem[OAG(2021)]{ccpa2021}
CA~OAG.
\newblock Ccpa regulations: Final regulation text.
\newblock \emph{Office of the Attorney General, California Department of
  Justice}, 2021.

\bibitem[Pawelczyk et~al.(2020{\natexlab{a}})Pawelczyk, Broelemann, and
  Kasneci]{pawelczyk2019}
Martin Pawelczyk, Klaus Broelemann, and Gjergji Kasneci.
\newblock Learning model-agnostic counterfactual explanations for tabular data.
\newblock In \emph{Proceedings of The Web Conference 2020 (WWW)}. ACM,
  2020{\natexlab{a}}.

\bibitem[Pawelczyk et~al.(2020{\natexlab{b}})Pawelczyk, Broelemann, and
  Kasneci]{pawelczyk2020multiplicity}
Martin Pawelczyk, Klaus Broelemann, and Gjergji. Kasneci.
\newblock On counterfactual explanations under predictive multiplicity.
\newblock In \emph{Proceedings of the 36th Conference on Uncertainty in
  Artificial Intelligence (UAI)}, pp.\  809--818. PMLR, 2020{\natexlab{b}}.

\bibitem[Pawelczyk et~al.(2021)Pawelczyk, Bielawski, Van~den Heuvel, Richter,
  and Kasneci]{pawelczyk2021carla}
Martin Pawelczyk, Sascha Bielawski, Johan Van~den Heuvel, Tobias Richter, and
  Gjergji Kasneci.
\newblock Carla: A python library to benchmark algorithmic recourse and
  counterfactual explanation algorithms.
\newblock In \emph{Advances in Neural Information Processing Systems (NeurIPS)
  (Benchmark and Datasets Track)}, 2021.

\bibitem[Pawelczyk et~al.(2022)Pawelczyk, Agarwal, Joshi, Upadhyay, and
  Lakkaraju]{pawelczyk2021connections}
Martin Pawelczyk, Chirag Agarwal, Shalmali Joshi, Sohini Upadhyay, and
  Himabindu Lakkaraju.
\newblock Exploring counterfactual explanations through the lens ofadversarial
  examples: A theoretical and empirical analysis.
\newblock In \emph{International Conference on Artificial Intelligence and
  Statistics (AISTATS)}, 2022.

\bibitem[Pawelczyk et~al.(2023)Pawelczyk, Datta, van-den Heuvel, Kasneci, and
  Lakkaraju]{pawelczyk2022noisy}
Martin Pawelczyk, Teresa Datta, Johannes van-den Heuvel, Gjergji Kasneci, and
  Himabindu Lakkaraju.
\newblock Algorithmic recourse in the face of noisy human responses.
\newblock In \emph{International Conference on Learning Representations
  (ICLR)}, 2023.

\bibitem[Rawal \& Lakkaraju(2020)Rawal and Lakkaraju]{rawal2020interpretable}
Kaivalya Rawal and Himabindu Lakkaraju.
\newblock Interpretable and interactive summaries ofactionable recourses.
\newblock In \emph{Advances in Neural Information Processing Systems
  (NeurIPS)}, volume~33, 2020.

\bibitem[Rawal et~al.(2021)Rawal, Kamar, and Lakkaraju]{rawal2021modelshifts}
Kaivalya Rawal, Ece Kamar, and Himabindu Lakkaraju.
\newblock Algorithmic recourse in the wild: Understanding the impact of data
  and model shifts.
\newblock \emph{arXiv:2012.11788}, 2021.

\bibitem[Rong et~al.(2022)Rong, Leemann, Borisov, Kasneci, and
  Kasneci]{rong2022evaluating}
Yao Rong, Tobias Leemann, Vadim Borisov, Gjergji Kasneci, and Enkelejda
  Kasneci.
\newblock A consistent and efficient evaluation strategy for attribution
  methods.
\newblock In \emph{International Conference on Machine Learning (ICML)}, 2022.

\bibitem[Shanmugam et~al.(2022)Shanmugam, Diaz, Shabanian, Finck, and
  Biega]{shanmugam2022dm}
Divya Shanmugam, Fernando Diaz, Samira Shabanian, Mich{\`e}le Finck, and
  Asia~J. Biega.
\newblock Learning to limit data collection via scaling laws: A computational
  interpretation for the legal principle of data minimization.
\newblock In \emph{ACM FAccT ’22}, 2022.

\bibitem[Slack et~al.(2021)Slack, Hilgard, Lakkaraju, and
  Singh]{slack2021counterfactual}
Dylan Slack, Sophie Hilgard, Himabindu Lakkaraju, and Sameer Singh.
\newblock Counterfactual explanations can be manipulated.
\newblock In \emph{Advances in Neural Information Processing Systems
  (NeurIPS)}, volume~34, 2021.

\bibitem[Spooner et~al.(2021)Spooner, Dervovic, Long, Shepard, Chen, and
  Magazzeni]{spooner2021counterfactual}
Thomas Spooner, Danial Dervovic, Jason Long, Jon Shepard, Jiahao Chen, and
  Daniele Magazzeni.
\newblock Counterfactual explanations for arbitrary regression models.
\newblock \emph{arXiv preprint arXiv:2106.15212}, 2021.

\bibitem[Strack et~al.(2014)Strack, DeShazo, Gennings, Olmo, Ventura, Cios, and
  Clore]{strack2014impact}
Beata Strack, Jonathan~P DeShazo, Chris Gennings, Juan~L Olmo, Sebastian
  Ventura, Krzysztof~J Cios, and John~N Clore.
\newblock Impact of hba1c measurement on hospital readmission rates: analysis
  of 70,000 clinical database patient records.
\newblock \emph{BioMed research international}, 2014, 2014.

\bibitem[Tolomei et~al.(2017)Tolomei, Silvestri, Haines, and
  Lalmas]{tolomei2017interpretable}
Gabriele Tolomei, Fabrizio Silvestri, Andrew Haines, and Mounia Lalmas.
\newblock Interpretable predictions of tree-based ensembles via actionable
  feature tweaking.
\newblock In \emph{Proceedings of the ACM SIGKDD International Conference on
  Knowledge Discovery \& Data Mining (KDD)}. ACM, 2017.

\bibitem[Union(2016)]{regulation2016regulation}
European Union.
\newblock Regulation (eu) 2016/679 of the european parliament and of the
  council.
\newblock \emph{Official Journal of the European Union}, 2016.

\bibitem[Upadhyay et~al.(2021)Upadhyay, Joshi, and
  Lakkaraju]{upadhyay2021robust}
Sohini Upadhyay, Shalmali Joshi, and Himabindu Lakkaraju.
\newblock Towards robust and reliable algorithmic recourse.
\newblock In \emph{Advances in Neural Information Processing Systems
  (NeurIPS)}, volume~34, 2021.

\bibitem[Ustun et~al.(2019)Ustun, Spangher, and Liu]{Ustun2019ActionableRI}
Berk Ustun, Alexander Spangher, and Y.~Liu.
\newblock Actionable recourse in linear classification.
\newblock In \emph{Proceedings of the Conference on Fairness, Accountability,
  and Transparency (FAT*)}, 2019.

\bibitem[Van~Looveren \& Klaise(2019)Van~Looveren and
  Klaise]{van2019interpretable}
Arnaud Van~Looveren and Janis Klaise.
\newblock Interpretable counterfactual explanations guided by prototypes.
\newblock \emph{arXiv preprint arXiv:1907.02584}, 2019.

\bibitem[Villaronga et~al.(2018)Villaronga, Kieseberg, and
  Li]{villaronga2018humans}
Eduard~Fosch Villaronga, Peter Kieseberg, and Tiffany Li.
\newblock Humans forget, machines remember: Artificial intelligence and the
  right to be forgotten.
\newblock \emph{Computer Law \& Security Review}, 34\penalty0 (2):\penalty0
  304--313, 2018.

\bibitem[Voigt \& Von~dem Bussche(2017)Voigt and Von~dem Bussche]{voigt2017eu}
Paul Voigt and Axel Von~dem Bussche.
\newblock The eu general data protection regulation (gdpr).
\newblock \emph{A Practical Guide, 1st Ed., Cham: Springer International
  Publishing}, 10:\penalty0 3152676, 2017.

\bibitem[Wachter et~al.(2018)Wachter, Mittelstadt, and
  Russell]{wachter2017counterfactual}
Sandra Wachter, Brent Mittelstadt, and Chris Russell.
\newblock Counterfactual explanations without opening the black box: automated
  decisions and the gdpr.
\newblock \emph{Harvard Journal of Law \& Technology}, 31\penalty0 (2), 2018.

\bibitem[Wightman(1998)]{wightman1998lsac}
Linda~F Wightman.
\newblock Lsac national longitudinal bar passage study. lsac research report
  series.
\newblock 1998.

\bibitem[Wu et~al.(2020)Wu, Dobriban, and Davidson]{wu2020deltagrad}
Yinjun Wu, Edgar Dobriban, and Susan Davidson.
\newblock Deltagrad: Rapid retraining of machine learning models.
\newblock In \emph{International Conference on Machine Learning (ICML)}, pp.\
  10355--10366. PMLR, 2020.

\bibitem[Xie et~al.(2017)Xie, Liang, and Song]{xie2017diverse}
Bo~Xie, Yingyu Liang, and Le~Song.
\newblock Diverse neural network learns true target functions.
\newblock In \emph{Artificial Intelligence and Statistics (AISTATS)}, pp.\
  1216--1224. PMLR, 2017.

\bibitem[Yamada et~al.(2020)Yamada, Lindenbaum, Negahban, and
  Kluger]{yamada20a}
Yutaro Yamada, Ofir Lindenbaum, Sahand Negahban, and Yuval Kluger.
\newblock Feature selection using stochastic gates.
\newblock In \emph{Proceedings of the 37th International Conference on Machine
  Learning (ICML)}, volume 119, 13--18 Jul 2020.

\bibitem[Zhang \& Zhang(2021)Zhang and Zhang]{zhang2021rethinking}
Rui Zhang and Shihua Zhang.
\newblock Rethinking influence functions of neural networks in the
  over-parameterized regime.
\newblock In \emph{Proceedings of the AAAI Conference on Artificial
  Intelligence (AAAI)}, 2021.

\end{thebibliography}
\bibliographystyle{iclr2023_conference}

\newpage
\appendix
\section*{Appendix}
\section{Theoretical Results}\label{appendix:theory}
\subsection{Upper Bounds on Recourse Outcome Instability}
\begin{innercustomprop}[Upper Bound on Output Robustness for Linear Models]
For the linear regression model $f(\bx) = \bw^\top \bx$ with weights given by $\bw=(\bX^\top \bX)^{-1} \bX^\top \bY$, an upper bound for the output robustness by removing an instance $(\bx, y)$ from the training set is given by:
\begin{align}
\Delta_\bx \leq \max_{i \in [n]}  ~ \lVert \bd_{i} \rVert_2 \cdot \lVert \xc_{\one} \rVert_2,
\end{align}
where $\bd_i = (\bX^\top \bX)^{-1} \bx_{i} \cdot \frac{r_i}{1 - h_{ii}}$, $r_i = y_i - \bw^\top \bx_i$ and $h_{ii} = \bx_i^\top (\bX^\top \bX)^{-1} \bx_i$.
\end{innercustomprop}

\begin{proof}
By Definition \ref{definition:ce_vulnerability_f}, we have:
\begin{align}
\Delta_\bx &= \big\lvert \bw_{\one}^\top \xc_{\one} - \bw_{-i}^\top \xc_{\one} \big \rvert \\
&= \big\lvert \big(\bw_{\one} - \bw_{-i} \big)^\top \xc_{\one} \big\rvert \notag \\
&= \bigg\lvert \bigg( (\bX^T \bX)^{-1} \bx_i \frac{(y_i - \bw^T \bx_i)}{1 - h_{ii} } \bigg)^\top \xc_{\one} \bigg\rvert && (\text{by Theorem \ref{theorem:leave_one_out_w}}) \\
& \leq \lVert \bd_i \rVert_2 \cdot  \lVert \xc_{\one} \rVert_2 && (\text{by Cauchy-Schwartz})\\
& \leq  \lVert \xc_{\one} \rVert_2 \cdot \max_{i \in [n]} ~ \lVert \bd_i \rVert_2 \cdot,
\end{align}
where $\bd_i = (\bX^T \bX)^{-1} \bx_i \frac{(y_i - \bw^T \bx_i)}{1 - h_{ii}}$.
This completes our proof.
\end{proof}


\begin{innercustomprop}[Upper Bound on Output Robustness for NTK]
For the NTK model with $\bw_{\text{NTK}} = \big(\bK^\infty(\bX, \bX)  + \lambda \bI_n \big)^{-1} \bY$, an upper bound for the output robustness by removing an instance $(\bx, y)$ from the training set is given by:
\begin{align}
\Delta_\bx \leq \lVert \bK^\infty(\xc_{\one}, \bX) \rVert_2 \cdot \max_{i \in [n]}  ~ \lVert \bd_{i} \rVert_2,
\end{align}
where $\bd_i = \frac{1}{k_{ii}} \bk_i \bk_i^\top \bY$, where $\bk_i$ is the $i$-th column of the matrix $\big(\bK^\infty(\bX, \bX)  + \beta \bI_n \big)^{-1}$, and $k_{ii}$ is its $i$-th diagonal element.
\end{innercustomprop}

\begin{proof}
By Definition \ref{definition:ce_vulnerability_f} and the weight-update theorem by \cite{zhang2021rethinking} (see Appendix \ref{app:prelim_theory}) and the assumption of the over-parameterized regime, we have:
\begin{align}
\Delta_\bx &= \big\lvert f_{\text{NTK}}(\xc_{\one}) - f^{-i}_{\text{NTK}}(\xc_{\one}) \big\rvert \notag \\
&= \big\lvert \big(\bK^\infty(\xc_{\one}, \bX) \big)^\top \bw_{\text{NTK}} - \big(\bK^\infty(\xc_{\one}, \bX) \big)^\top \bigg((\bK^\infty(\bX, \bX) + \beta \bI_n \big)^{-1} - \frac{1}{k_{ii}} \bk_i \bk_i^\top \bigg) \bY \big\rvert \\
& = \big\lvert \big(\bK^\infty(\xc_{\one}, \bX) \big)^\top \frac{1}{k_{ii}} \bk_i \bk_i^\top \bY \big\rvert \\ 
& \leq \lVert \bd_i \rVert_2 \cdot  \lVert \bK^\infty(\xc_{\one}, \bX)  \rVert_2 ~~~ (\text{by Cauchy-Schwartz}) \notag\\
& \leq \lVert \xc_{\one} \rVert_2 \cdot \max_{i \in [n]} ~ \lVert \bd_i \rVert_2,
\end{align}
where $\bd_i=\frac{1}{k_{ii}} \bk_i \bk_i^\top \bY$ which completes our proof.
\end{proof}

\subsection{Upper Bounds on Recourse Action Instability}
\begin{innercustomprop}[Upper Bound on Input Robustness]
For the linear regression model $f(\bx) = \bw^\top \bx$ with weights given by $\bw=(\bX^\top \bX)^{-1} \bX^\top \bY$, an upper bound for the input robustness in the setting $s=0, \lambda=0$ by removing the $i$-th instance $(\bx_i, y_i)$ from the training set is given by:
\begin{align}
\Phi^{(2)}_\bx \leq \lVert \bd_i \rVert_2 \frac{4\sqrt{2} \lVert \bx \rVert_2}{\min (\lVert \bw \rVert_2, \lVert \bw_{-i} \rVert_2)},
\end{align}
under the condition that $\bw^\top\bw_{-i}\leq 0$ (no diametrical weight changes), where $\bw_{-i}=\bw-\bd_i$ is the weight after removal of training instance $i$ and $\bd_i = (\bX^T \bX)^{-1} \bx_i \frac{(y_i - \bw^\top \bx_i)}{1 - h_{ii}}$.
\end{innercustomprop}

\begin{proof}
For a linear scoring function $f(\bx)=\bw'^\top \bx$ with given parameters $\bw'$, under the squared $\ell_2$ norm constraint with balance parameter $\lambda$, the optimal recourse action is given by \citep{pawelczyk2021connections}:
\begin{align}
\bdelta\left(\bw'\right) = \frac{s-\bw'^\top\bx}{\lVert \bw' \rVert_2^2 + \lambda} \cdot \bw'.
\end{align}
Using Definition \ref{definition:ce_vulnerability_x}, we can express the total change in $\bdelta$ as a path integral over changes in $\bw$, times the change $\frac{\bD\bdelta}{\bD\bw}$ they entail:
\begin{align}
\Phi^{(2)}_\bx &= \big\lVert \bdelta_{\one} - \bdelta_{\bomega}\big\rVert_2 = \big\lVert \bdelta\left(\bw \right) - \bdelta\left(\bw_{-i} \right)\big\rVert_2 \\
&\leq \int_0^1 \bigg\lVert \frac{\bD\bdelta}{\bD\bw} \left(\gamma\bw +(1-\gamma)\bw_{-i}\right) \bigg\rVert \lVert\bw-\bw_{-i}\rVert_2 d\gamma,
\end{align}
where $\frac{\bD\bdelta}{\bD\bw}$ denotes the Jacobian, with the corresponding operator matrix norm. Defining $\tilde{\bw} \coloneqq \gamma\bw +\left(1-\gamma\right)\bw_{-i}$ and using $\lVert\bw-\bw_{-i}\rVert_2=\lVert \bd_i \rVert_2$, we obtain
\begin{align}
\Phi^{(2)}_\bx &\leq \lVert \bd_i \rVert_2 \int_0^1 \bigg\lVert \frac{\bD\bdelta}{\bD\bw}\left(\tilde{\bw} \right)\bigg\rVert_2d\gamma.
\end{align}
Because of the form $\bdelta(\bw')=f(\bw')\bw'$, where $f(\bw') \coloneqq \frac{s-\bw'^\top\bx}{\lVert \bw' \rVert_2^2 + \lambda}$ is a scalar function, its Jacobian has the form $\frac{\bD\bdelta}{\bD\bw'} = \bw'\left(\nabla f(\bw')\right)^\top  + f(\bw')\bI$.
We will now derive a bound on the Jacobian's operator norm:
\begin{align}
    \bigg\lVert\frac{\bD\bdelta}{\bD\bw'}\left(\tilde{\bw} \right)\bigg\rVert_2 &= \max_{\lVert \ba \rVert = 1} \bigg\lVert \frac{\bD\bdelta}{\bD\bw'} \ba \bigg\rVert_2 =  \max_{\lVert \ba \rVert = 1} \bigg\lVert \bw'\left(\nabla f(\tilde{\bw})\right)^\top \ba   + f(\tilde{\bw}) \ba \bigg\rVert_2\\
    & \leq \lVert \nabla f(\tilde{\bw}) \rVert_2 \lVert \tilde{\bw} \rVert_2 +\lvert f(\tilde{\bw}) \rvert.
\end{align}
Additionally, we know that for $s=0$, $\lvert f(\bw') \rvert \leq \frac{\lVert \bx \rVert_2 \lVert \tilde{\bw} \rVert_2}{\lVert \tilde{\bw} \rVert_2^2}=\frac{\lVert \bx \rVert_2}{\lVert \tilde{\bw} \rVert_2}$. The gradient is given by
\begin{align}
    \lVert \nabla f(\tilde{\bw}) \rVert_2 &= \bigg \lVert\frac{-(\lVert \tilde{\bw} \rVert_2^2 + \lambda)\bx - 2(s-\tilde{\bw}^\top\bx)\tilde{\bw}}{(\lVert \tilde{\bw} \rVert_2^2 + \lambda)^2} \bigg \rVert_2 \\
    &\leq \frac{(\lVert \tilde{\bw} \rVert_2^2 + \lambda)\lVert \bx \rVert_2 + 2(s+\lVert \tilde{\bw} \rVert_2\lVert \bx \rVert_2)\lVert_2\tilde{\bw} \rVert_2}{\lVert \tilde{\bw} \rVert_2^4}\\
    &= \frac{3\lVert \bx \rVert_2}{\lVert \tilde{\bw} \rVert_2^2}&& (\text{Using $\lambda \to 0$, $s=0$}).
\end{align}
In summary,
\begin{align}
    \bigg\lVert\frac{\bD\bdelta}{\bD\bw'}\left(\tilde{\bw} \right)\bigg\rVert_2  \leq \frac{3\lVert \bx \rVert_2}{\lVert \tilde{\bw} \rVert_2^2} \lVert \tilde{\bw} \rVert_2 + \frac{\lVert \bx \rVert_2}{\lVert \tilde{\bw} \rVert_2} = \frac{4\lVert \bx \rVert_2}{\lVert \tilde{\bw} \rVert_2}.
\end{align}
Because $\tilde{\bw}$ is a line between $\bw$ and $\bw_{-i}$, its norm is bounded from below by $\lVert \tilde{\bw} \rVert_2 \geq \frac{1}{\sqrt{2}}\min (\lVert \bw \rVert_2, \lVert \bw_{-i} \rVert_2) \geq \frac{1}{\sqrt{2}}\left(\lVert \bw \rVert_2 -\lVert\bw-\bw_{-i}\rVert_2\right)= \frac{1}{\sqrt{2}}\left(\lVert \bw \rVert_2 - \lVert \bd_i \rVert_2 \right)$. We can thus uniformly bound the integral and plug in the bound because of its positivity, 
\begin{align}
\Phi^{(2)}_\bx &\leq \lVert \bd_i \rVert_2 \int_0^1 \bigg\lVert \frac{\bD\bdelta}{\bD\bw}\left(\tilde{\bw} \right)\bigg\rVert_2d\gamma \\
&\leq \lVert \bd_i \rVert_2 \int_0^1 \frac{4\sqrt{2}\lVert \bx \rVert_2}{\min (\lVert \bw \rVert_2, \lVert \bw_{-i} \rVert_2)} d\gamma \\
& = \lVert \bd_i \rVert_2 \frac{4\sqrt{2} \lVert \bx \rVert_2}{\min (\lVert \bw \rVert_2, \lVert \bw_{-i} \rVert_2)},
\end{align}

which completes the proof.
\end{proof}

\subsection{Calculating Recourse Outcome Instability for k Deletions is NP-hard}
\label{app:np_hardness}
We can show that, for a general scoring function $f$, the problem defined in \eqref{eq:optimize_invalidation} is NP-hard. We make this proof by providing a function $f$ for which solving the recourse outcome invalidity problem is as hard as solving the well-known Knapsack problem, that has been shown to be NP-hard \citep{karp1972reducibility}. The knapsack problem is defined as follows:
\begin{align}
    \max_{q_i \in \{0,1\}}  \sum_{i=1}^n v_i q_i 
    \text{  s.t.  } & \sum_{i=1}^n y_i q_i \leq W,
\end{align}
where the problem considers $n$ fixed items $(v_i, y_i)_{i=1\ldots n}$ with a value $v_i$ and knapsack weight $y_i > 0$, and $W$ is a fixed weight budget.
The optimization problem consists of choosing the items that maximize the summed values but have a weight lower than $W$. 
To solve this problem through the recourse outcome invalidation problem, we suppose there is a data point for each item. We can choose any $ k > \frac{W}{\min{y_i}}$ of points to be deleted, where this condition ensures that we can remove the number of samples maximally required to solve the corresponding knapsack problem. Note that we can always add a number of dummy points that have no effect such that the total number of data points is at least $k$. Suppose there is a classifier function:
\begin{align}
    f_{\bomega}(\bx) \coloneqq \left\{\begin{array}{ll}
        \sum_{i=1}^{n} v_i (1-\omega_i),& \sum_{i=1}^{n} y_i (1-\omega_i) \leq W\\
        0, &\text{else} \\\end{array}\right. .
\end{align}
In this case, solving Eqn. \ref{eq:optimize_invalidation} comes down to finding the set of items (i.e., removing the data points) that have maximum value, but stay under the threshold $W$. Thus, if we can solve Eqn. \ref{eq:optimize_invalidation}, the solution to the equivalent knapsack problem is given by $\mathbf{q}=(\mathbf{1}-\bomega)$. 


\subsection{Auxiliary Theoretical Results}
\label{app:prelim_theory}
We state the following classic result by \cite{miller1974unbalanced} without proof.
\begin{theorem}(Leave-One-Out Estimator, \cite{miller1974unbalanced})\label{theorem:leave_one_out_w}
Define $(\bx_i, y_i)$ as the point to be removed from the training set.
Given the optimal weight vector $\bw=(\bX^\top \bX)^{-1} \bX^\top \bY$ which solves for a linear model under mean-squared-error loss, the leave-one-out estimator is given by:
\begin{equation*}
    \bw - \bw_{-i} = (\bX^T \bX)^{-1} \bx_i \frac{(y_i - \bw^T \bx_i)}{1 - \bx_i^T (\bX^T \bX)^{-1} \bx_i } = (\bX^T \bX)^{-1} \bx_i \frac{(y_i - \bw^T \bx_i)}{1 - h_{ii} } =: \bd_i.
\end{equation*}
\end{theorem}
\replace{
We restate the analtical solution for the NTK weights in case of a single deletion from \cite{zhang2021rethinking}.
\begin{theorem}(Leave-One-Out weights for NTK models, \cite{zhang2021rethinking})\label{theorem:zhang}
Let $\bw_{\text{NTK}} = \big(\bK^\infty(\bX, \bX)  + \lambda \bI_n \big)^{-1} \bY$ be the weight for the NTK model on the full data Kernel model, where $\bK^\infty(\bX, \bX)$ is the NTK matrix evaluated on the training data points: $[\bK^\infty(\bX, \bX) ]_{ij}=K^\infty(\bx_i, \bx_j)$. Then, the NTK model that would be obtained when removing instance $i$, could be equivalently described by
\begin{equation*}
    f_{\text{NTK}}^{-i}\left(\bx\right) =\bK^\infty(\bx, \bX)^\top \bw_{\text{NTK},-i}= \bK^\infty(\bx, \bX)^\top\left(\big(\bK^\infty(\bX, \bX)  + \lambda \bI_n \big)^{-1} -\frac{1}{q_{-ii}} \vq_{-i} \vq_{-i}^\top \right)\bY
\end{equation*}
where $\vq_{-i}$ is the $i$-th column of the matrix $\mQ^{-1} =\big(\bK^\infty(\bX, \bX)  + \beta \bI_n \big)^{-1}$, and $q_{-ii}$ is its $i$-th diagonal element of this inverse.
\end{theorem}

\begin{proof}
We begin by introducing some notation. Define: 
 \begin{align}
     \mQ &= \bK^{\infty} (\bX,\bX) + \beta \bI_n \\
     \mR &=\bK^{\infty} (\bX_{-i},\bX_{-i} )+ \beta \bI_{n-1},
 \end{align} 
 then the analytical NTK model when considering the dataset with one instance $\bx_i$ removed is given by
 \begin{align}
 \label{eqn:smallerntk}
 f_{\text{NTK}}^{-i}\left(\bx\right) = \bK^\infty(\bx, \bX_{-i})^\top \mR^{-1} \bY_{-i},
 \end{align}
where $\bX_{-i}$ denotes the data matrix with row $i$ missing and $\bY_{-i}$ denotes the label vector with the $i$-th label missing. We have to show that this expression is equivalent to that stated in the theorem.

Without loss of generality, we can assume the $i$ is the last point in the dataset (otherwise, we just permute the data set accordingly).
Therefore, we can write the matrix $\mQ$ in block form:
\begin{align}
    \mQ=\begin{bmatrix} \mR & \bK^\infty(\bx_i, \mX) \\
    \bK^{\infty\top}(\bx_i, \mX) & \bK^\infty(\bx_i, \bx_i) \\
    \end{bmatrix} \coloneqq \begin{bmatrix} \mR & \vq_i \\
    \vq_i^\top & q_{ii}\\
    \end{bmatrix}
\end{align}
Through the block matrix inversion formula (see for example \cite{csato2002sparse} (eqn.\ 52)) we can write $\mQ$'s inverse as
\begin{align}
    \mQ^{-1}=\begin{bmatrix} \mR^{-1}+\gamma^{-1}\mR^{-1} \vq_i \vq_i^\top \mR^{-1} & \gamma^{-1} \mR^{-1}\vq_i \\
    \gamma^{-1}\vq_i^\top \mR^{-1} & \gamma^{-1} \\
    \end{bmatrix}
\label{eq:q_inverse}
\end{align}
with $\gamma=q_{ii}-\vq_i^\top \mR^{-1} \vq_i$.

We denote the $i$-th (and last) column of $\mQ^{-1}$ as $\vq_{-i}=\begin{bmatrix}\gamma^{-1} \mR^{-1}\vq_i\\ \gamma^{-1}\end{bmatrix}$
and the $i$-th and last diagonal element of the inverse as $q_{-ii}=\gamma^{-1}$.
We will now show, that the form of the weights given in the theorem (i.e., the weights for the points not removed) are equivalent to the weights that would have been computed by plugging in the smaller kernel matrix $\bK^\infty(\bX_{-i}, \bX_{-i})$ in the analytical solution and the weight for the point deleted will have a value of zero, i.e.,
\begin{align}
\left(\big(\bK^\infty(\bX, \bX)  + \beta \bI_n \big)^{-1} -\frac{1}{q_{-ii}} \vq_{-i} \vq_{-i}^\top \right)\bY & = \left(\mQ^{-1} -\frac{1}{q_{-ii}} \vq_{-i} \vq_{-i}^\top \right)\bY
\label{eq:update} 
\\
& = \begin{bmatrix}\mR^{-1}\bY_{-i}\\0\end{bmatrix}.
\end{align}

To show this, we plug in the inversion formula $\mQ^{-1}$ from \eqref{eq:q_inverse} into \eqref{eq:update} and using $q_{-ii}=\gamma^{-1}$:
\begin{align}
    & \left(\mQ^{-1} -\frac{1}{q_{-ii}} \vq_{-i} \vq_{-i}^\top \right)\bY\\
    &=\left(\begin{bmatrix} \mR^{-1}+\mR^{-1} \vq_i \vq_i^\top \mR^{-1} & \gamma^{-1} \mR^{-1}\vq_i \\
    \gamma^{-1}\vq_i^\top \mR^{-1} & \gamma^{-1} \\
    \end{bmatrix}-\frac{1}{\gamma^{-1} }\begin{bmatrix}\gamma^{-1} \mR^{-1}\vq_i\\ \gamma^{-1}\end{bmatrix} \begin{bmatrix}\gamma^{-1} \mR^{-1}\vq_i\\ \gamma^{-1}\end{bmatrix}^\top \right)
    \begin{bmatrix}\bY_{-i}\\Y_i\end{bmatrix}\\
    &=  \left(\begin{bmatrix} \mR^{-1}+\gamma^{-1}\mR^{-1} \vq_i \vq_i^\top \mR^{-1} & \gamma^{-1} \mR^{-1}\vq_i \\
    \gamma^{-1}\vq_i^\top \mR^{-1} & \gamma^{-1} \\
    \end{bmatrix}-{\gamma} \begin{bmatrix} \gamma^{-2} \mR^{-1}\vq_i\vq_i^\top\mR^{-1} & \gamma^{-2} \mR^{-1}\vq_i \\
    \gamma^{-2} \vq_i^\top\mR^{-1} & \gamma^{-2} \\
    \end{bmatrix} \right)
    \begin{bmatrix}\bY_{-i}\\Y_i\end{bmatrix}\\
&= \begin{bmatrix} \mR^{-1}+ \gamma^{-1}\mR^{-1} \vq_i \vq_i^\top \mR^{-1} -\gamma^{-1} \mR^{-1}\vq_i\vq_i^\top\mR^{-1} & \gamma^{-1} \mR^{-1}\vq_i- \gamma^{-1} \mR^{-1}\vq_i\\
    \gamma^{-1}\vq_i^\top \mR^{-1} -\gamma^{-1} \vq_i^\top\mR^{-1} & \gamma^{-1} -\gamma^{-1}  \\
    \end{bmatrix}    \begin{bmatrix}\bY_{-i}\\Y_i\end{bmatrix}\\
    & = 
    \begin{bmatrix} \mR^{-1} & \mathbf{0} \\
    \mathbf{0}^\top & 0  \\
    \end{bmatrix}    \begin{bmatrix}\bY_{-i}\\Y_i\end{bmatrix} = \begin{bmatrix}\mR^{-1}\bY_{-i}\\0\end{bmatrix}.
\end{align}
Therefore, we have equivalence between \eqref{eqn:smallerntk} and the formulation in the theorem.

\end{proof}
}
\subsection{An Analytical NTK Kernel}
\label{sec:ntk}
In this section, we provide theoretical results that allow deriving the closed form solution of the NTK for the two-layer ReLU network. First, see the paper by Jacot et al. \cite{jacot2018neural} for the original derivation of the neural tangent kernel.

\textbf{A closed-form solution for two-layer ReLU networks.} From \cite[Assumption 3.1]{zhang2021rethinking, du2018gradient} we obtain the definition of the Kernel matrix $\bK^\infty$ (termed Gram matrix in the paper \cite{du2018gradient}) for ReLU networks:
\begin{align*}
\bK^\infty_{ij} = \bK^\infty(\bx_i, \bx_j) &= \mathbb{E}_{\bw\sim\mathcal{N}\left(0,\bI \right)}\left[\bx_i^\top\bx_j \mathbb{I}\left\{\bw^\top \bx_i \geq 0, \bw^\top \bx_j \geq 0\right\} \right] \\
&= \bx_i^\top\bx_j\mathbb{E}_{\bw\sim\mathcal{N}\left(0,\bI \right)}\left[ \mathbb{I}\left\{\bw^\top \bx_i \geq 0, \bw^\top \bx_j \geq 0\right\} \right] \\
&= \bx_i^\top\bx_j  \frac{\pi - \text{arcos}\left(\frac{\bx_i^\top \bx_j}{\lVert\bx_i\rVert\lVert\bx_j\rVert}\right)}{2 \pi}.
\end{align*}
The last reformulation uses an analytical result by \cite{cho2009kernel}. The derived result matches the one by \cite{xie2017diverse}, which however does not provide a comprehensive derivation.

\section{Additional Experimental Results}\label{appendix:exp_results}
\textbf{Data sets for the Classification Tasks}
When considering classification tasks on the \emph{heloc} and \emph{admission} data sets, we threshold the scores based on the median to obtain binary target labels.
On the Admission data set (in the classification setting), a counterfactual is found when the predicted first-year average score switches from ‘below median’ to ‘above median’. We then count an invalidation if, after the model update, the score of a counterfactual switches back to ‘below median’.
In addition to the aforementioned data sets, we use both the \emph{Diabetes} and the \emph{Compas} data sets.
The \emph{Diabetes} data set which contains information on diabetic patients from 130 different US hospitals \citep{strack2014impact}. 
The patients are described using administrative (e.g., length of stay) and medical records (e.g., test results), and the prediction task is concerned with identifying whether a patient will be readmitted within the next 30 days.
We sub sampled a smaller data sets of 10000 points from this dataset. 
8000 points are left to train the model, while 2000 points are left for the test set. 
The \emph{Compas} data set \cite{Angwin2016} contains data for more than 10,000 criminal defendants in Florida.
It is used by the jurisdiction to score defendant's likelihood of reoffending. 
We kept a small part of the raw data as features like \textit{name}, \textit{id}, \textit{casenumbers} or \emph{date-time} were dropped. 
The classification task consists of classifying an instance into high risk of recidivism.
Across all data sets, we dropped duplicate instances.

\begin{figure*}[htb]
\centering
\begin{subfigure}{\textwidth}
\centering
\includegraphics[width=\textwidth]{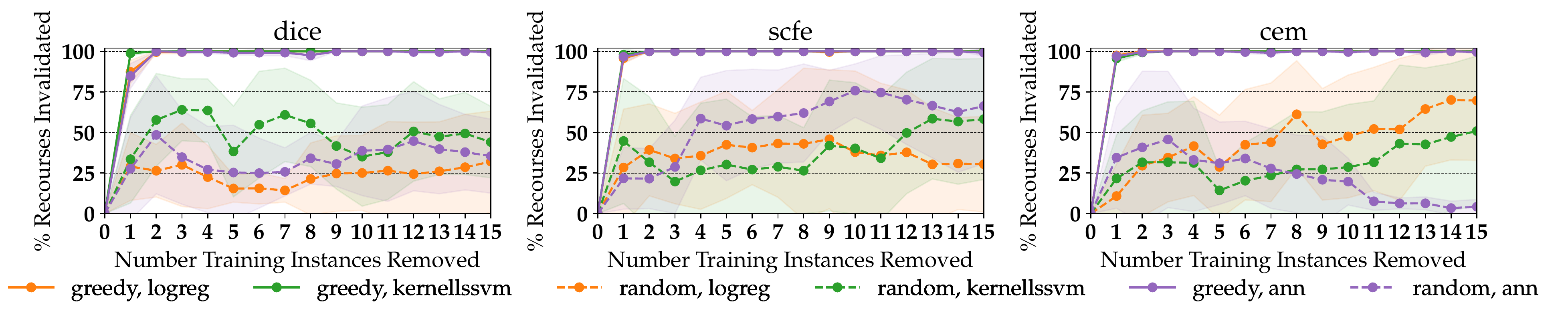}
\caption{Admission} 
\label{fig:invalidation_output_admission_greedy}
\end{subfigure}
\vfill 
\begin{subfigure}{\textwidth}
\centering
\includegraphics[width=\textwidth]{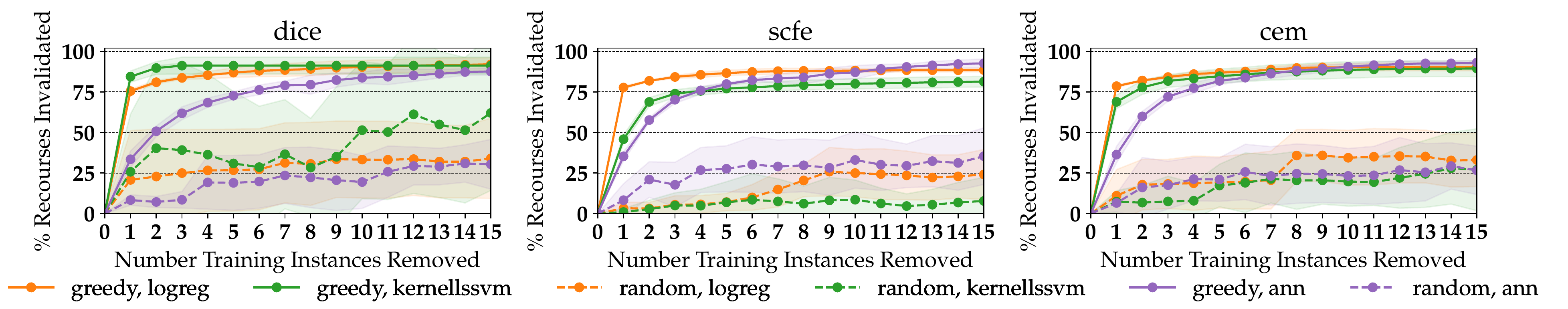}
\caption{Heloc} 
\label{fig:invalidation_output_heloc_greedy}
\end{subfigure}
\vfill 
\begin{subfigure}{\textwidth}
\centering
\includegraphics[width=\textwidth]{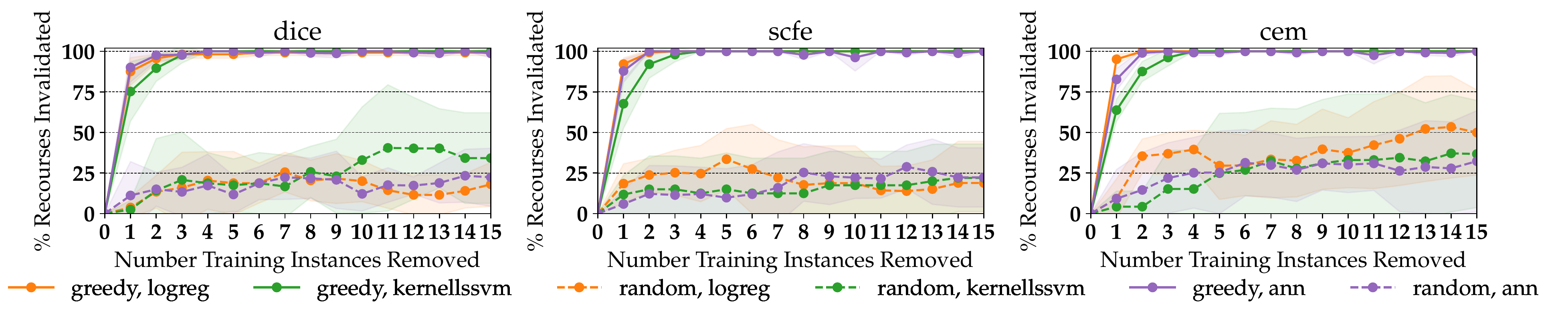}
\caption{Compas} 
\label{fig:invalidation_output_compas_greedy}
\end{subfigure}
\vfill 
\begin{subfigure}{\textwidth}
\centering
\includegraphics[width=\textwidth]{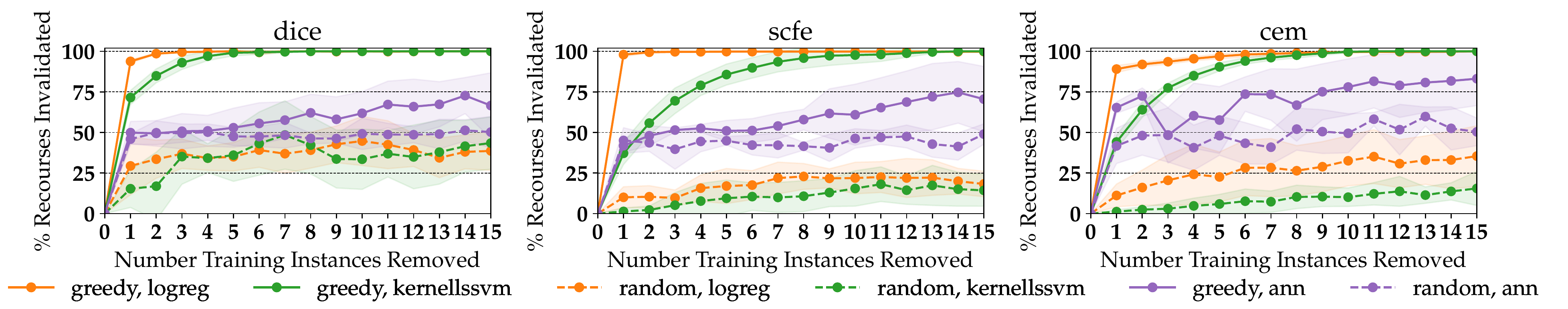}
\caption{Diabetes} 
\label{fig:invalidation_output_diabetes_greedy}
\end{subfigure}
\caption{Measuring the tradeoff between \recourseo instability and the number of \delrequests for the Admission, Heloc, Diabetes and Compas data sets for logistic regression, kernel svm, and ANN models across recourse methods on classification tasks. Results were obtained by greedy optimization. The dotted lines indicate the random baselines.}
\label{fig:invalidation_output_classification_greedy}
\end{figure*}

\textbf{Discussing the Results}
As suggested in Section \ref{section:experiments}, here we are discussing  the remaining \recourseo invalidation results.
We show these results for two settings. 
In Figure \ref{fig:invalidation_output_classification_greedy}, we demonstrate the efficacy of our greedy deletion algorithm across 4 data sets on the classification tasks using different classification models (ANN, logistic regression, Kernel-SVM). For the logistic regression and the ANN model, we use the infinitesimal jackknife approximation to calculate the probitively expensive retraining step as described in Section \ref{sec:optimizationalgorithms}.
We observe that our method well outperforms random guessing. The results also highlight that while the NTK theory allows to study the deletion effects from a theoretical point of view, if one is interested in empirical worst-case approximations, the infinitesimal jackknife can be a method of choice.
As we observe this pattern across all recourse methods, we hypothesize that this is related to the instability of the trained ANN models, and we leave an investigation of this interesting phenomenon for future work.

\begin{figure*}[htb]
\centering
\begin{subfigure}{\textwidth}
\centering
\includegraphics[width=\textwidth]{figures/invalidation_output_admission_regression_greedy.pdf}
\caption{Admission (Greedy)} 
\label{fig:invalidation_output_admission_greedy}
\end{subfigure}
\vfill 
\begin{subfigure}{\textwidth}
\centering
\includegraphics[width=\textwidth]{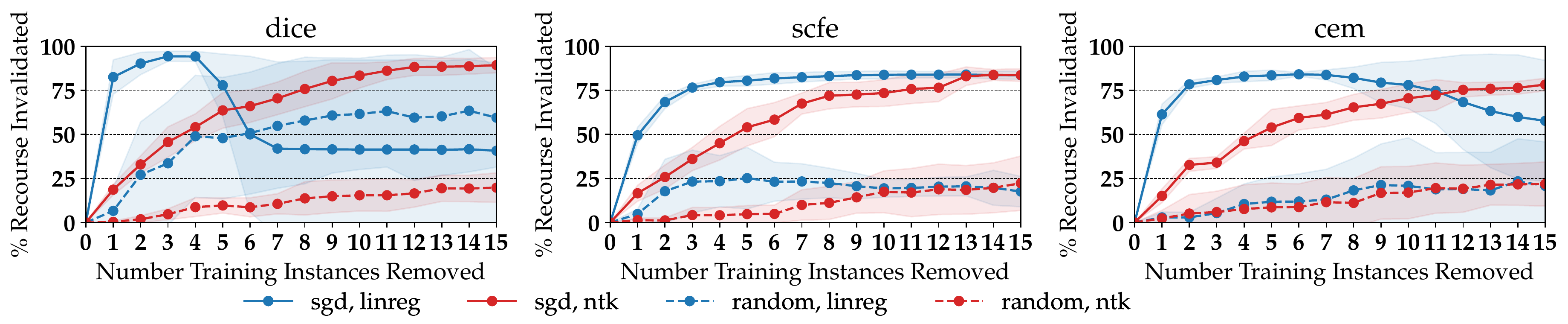}
\caption{Admission (SGD)} 
\label{fig:invalidation_output_admission_sgd}
\end{subfigure}
\vfill 
\begin{subfigure}{\textwidth}
\centering
\includegraphics[width=\textwidth]{figures/invalidation_output_heloc_regression_greedy.pdf}
\caption{Heloc (Greedy)} 
\label{fig:invalidation_output_heloc_greedy}
\end{subfigure}
\vfill 
\begin{subfigure}{\textwidth}
\centering
\includegraphics[width=\textwidth]{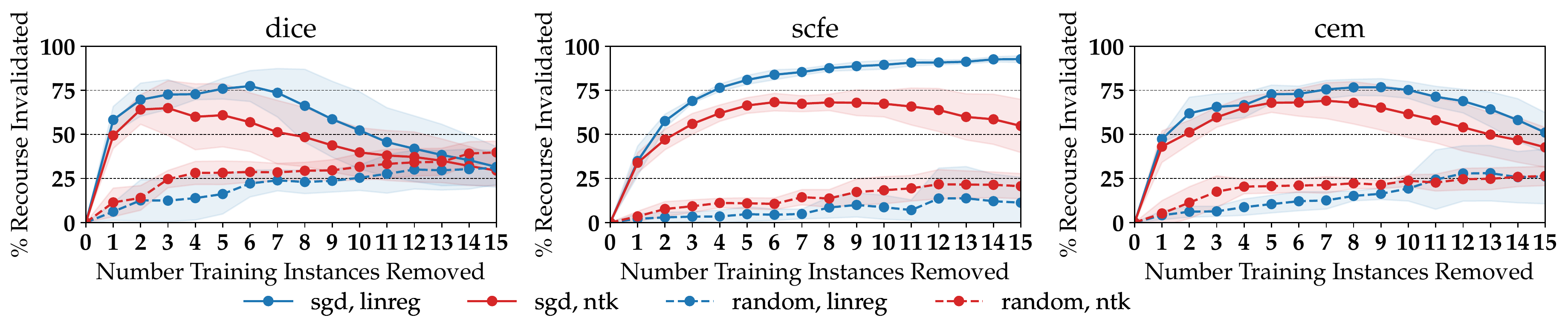}
\caption{Heloc (SGD)} 
\label{fig:invalidation_output_heloc_sgd}
\end{subfigure}
\caption{Measuring the tradeoff between \recourseo instability and the number of \delrequests for the Admission and Heloc data sets for linear regression and NTK models across recourse methods on regression tasks. Results were obtained by both SGD and Greedy optimization. The dotted lines indicate the random baselines.}
\label{fig:invalidation_output_regression_sgd}
\end{figure*}

Additionally, in Figure \ref{fig:invalidation_output_regression_sgd}, we compare our SGD-based deletion algorithm to the greedy algorithm. 
For the SGD-based deletion results, we observe inverse-u-shaped curves on some method-data-model combinations. 
The reason for this phenomenon can be explained as follows: when the $\ell_0$ regularization strength (i.e., $\eta$) is not strong enough, then the importance weights for the $k$-th removal with $k>5$ become more variable (i.e., SGD does not always select the most important data weight for larger $k$).
This drop in performance can be mitigated by increasing the strength of the $\ell_0$ regularizer within our SGD-based deletion algorithm.

\begin{figure*}[htb]
\centering
\begin{subfigure}{0.49\textwidth}
\centering
\includegraphics[width=\textwidth]{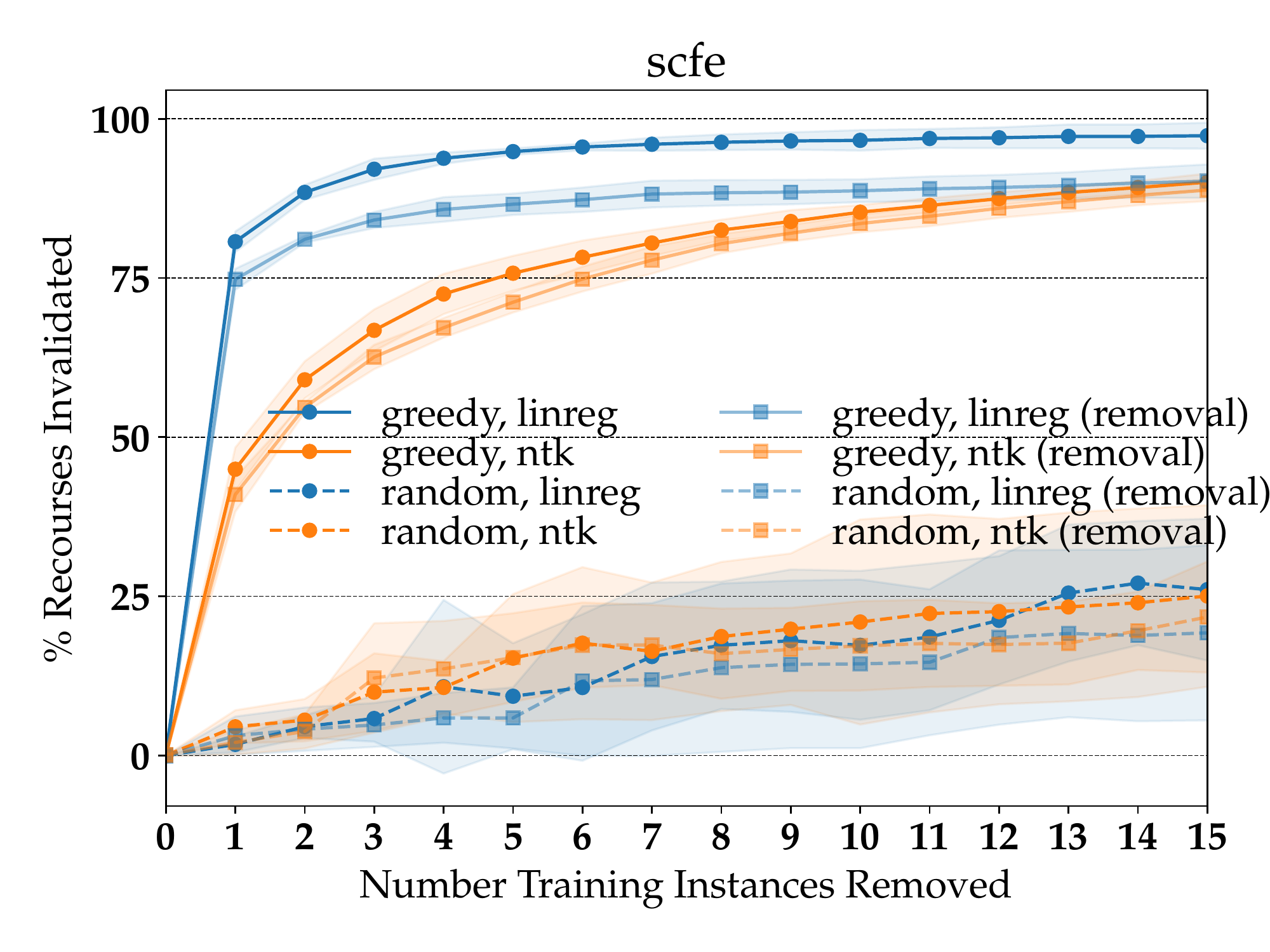}
\caption{Heloc (Greedy)} 
\end{subfigure}
\hfill
\begin{subfigure}{0.49\textwidth}
\centering
\includegraphics[width=\textwidth]{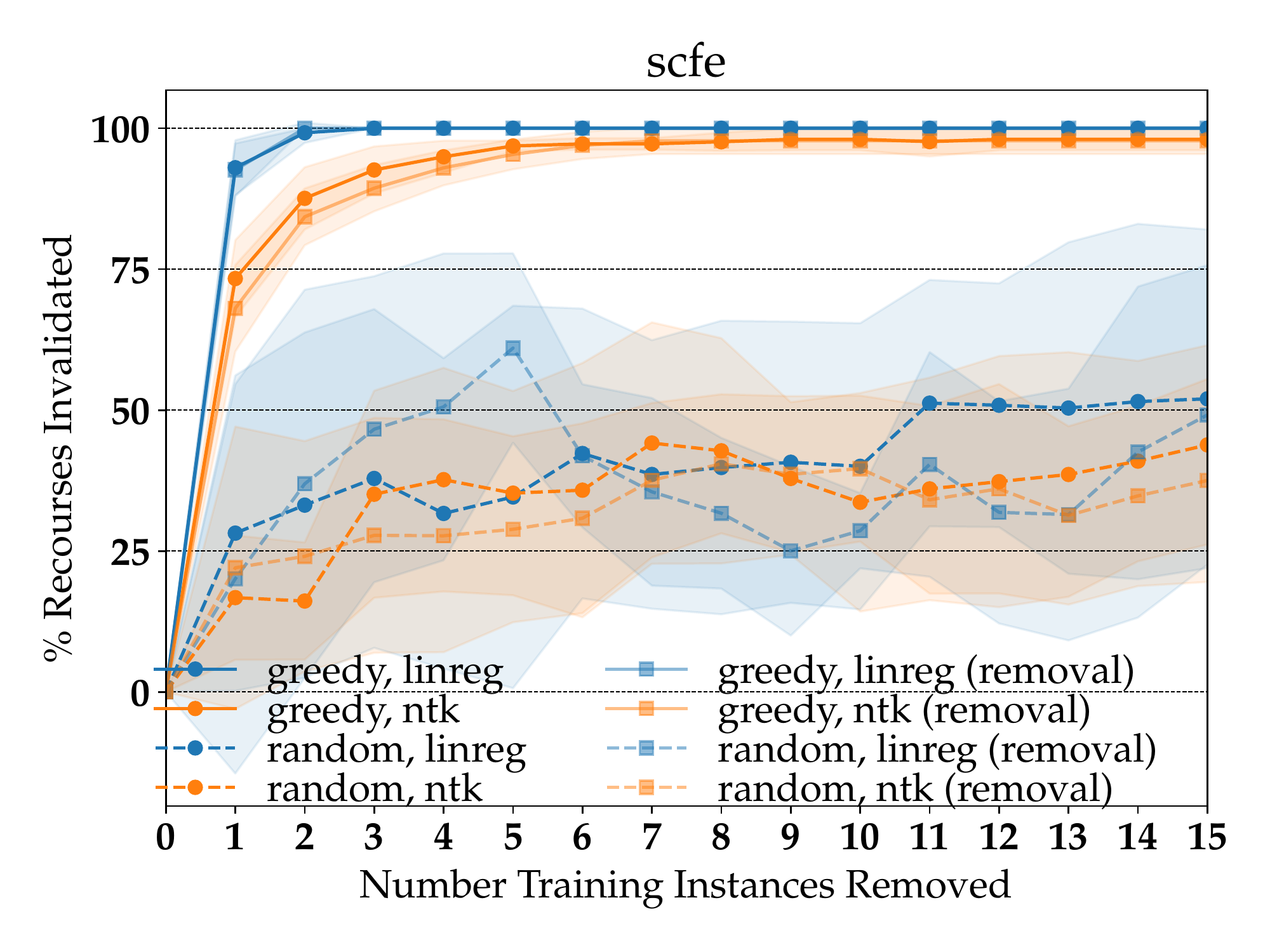}
\caption{Admission (Greedy)} 
\end{subfigure}
\caption{Measuring the efficacy of a simple removal strategy on the Heloc and Admission data set for linear and NTK regression models.
We removed the 15 critical points identified for the linear and NTK models when the underlying recourse method is \texttt{SCFE} and reran the removal algorithm on the remaining training set. Results were obtained by Greedy optimization. The dotted lines indicate the random baselines.}
\label{fig:onion_effect}
\end{figure*}

In Figure \ref{fig:onion_effect},  we study a simple removal strategy aimed at increasing the stability of algorithmic recourse. 
To this end, we identified the 15 points that lead to the highest invalidation on the NTK and linear regression models when the underlying recourse method is \texttt{SCFE}. 
Using our greedy method, we remove these 15 points from the training data set, and we then then rerun our proposed greedy removal algorithm. 
This strategy leads to an improvement of up to 6 percentage points over the initial model where the 15 most critical points were included, suggesting that the removal of these critical points can be used to alleviate the recourse instability issue. 
In future work, we plan to investigate strategies that increase the robustness of algorithmic recourse even further.

\begin{figure*}[htb]
\centering
\begin{subfigure}{0.5\textwidth}
\centering
\includegraphics[width=\textwidth]{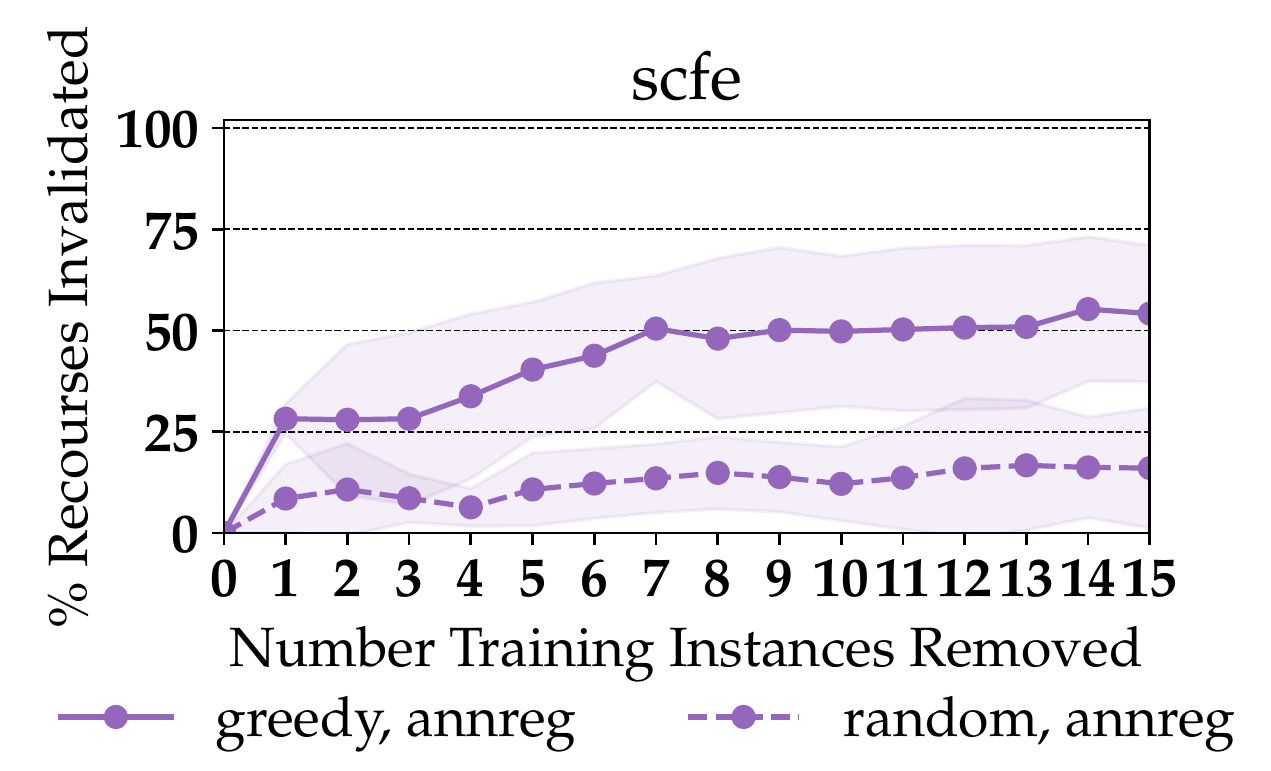}
\caption{Admission (Greedy)} 
\end{subfigure}
\caption{Measuring the tradeoff between \recourseo instability and the number of \delrequests for the Admission data set for a neural network regression model.
We used the critical points identified for the NTK model to invalidate the recourses identified by a wide 2-layer ReLU network with 10000 hidden nodes. Results were obtained by Greedy optimization. The dotted lines indicate the random baselines.}
\label{fig:ntk_approx}
\end{figure*}

Finally, in Figure \ref{fig:ntk_approx} we study how well the critical points identified for the NTK model would invalidate a wide 2-layer ReLU network with 10000 hidden nodes. To study that question, we identified the points that lead to the highest invalidation on the NTK using our greedy method, and we then use these identified training points to invalidate the recourses suggested by the wide ANN. As before, we are running these experiments on the full data set across 5 folds. Figure \ref{fig:ntk_approx} demonstrates the results of this strategy for the \texttt{SCFE} recourse method. We see that this strategy increases the robustness of up to 30 percentage points over the random baseline, suggesting that critical points under NTK can be used to estimate recourse invalidation for wide ANN models.

\section{Implementation Details}\label{appendix:implementaion_details}

\subsection{Details on Model Training}\label{appendix:ml_classifiers}
We train the classification models using the hyperparameters given in Table~\ref{tab:hyperparams}. The ANN and the Logistic regression models are fit using the quasi-newton \texttt{lbgfs} solver. We add L2-regularization to the ANN weights.
The other methods are trained via their analytical solutions.
Below, in Algorithms \ref{algorithm:greedy_deletion} and \ref{algorithm:sgd_deletion}, we show pseudocodes for both our greedy and sgd-based deletion methods to invalidate the \recourseo. 
In order to do the optimization with respect to the recourse action stability measure, we slightly adjust Algorithm \ref{algorithm:sgd_deletion} to optimize the right metric from Definition \ref{definition:ce_vulnerability_x}.

\subsection{Details on Generating the Counterfactuals}
For \texttt{DICE}, for every test input, we generate two different counterfactual explanations. Then we randomly pick either the first or second counterfactual to be the counterfactual assigned to the given input.
Across all recourse methods the success rates lie above 95\%, i.e., for 95\% of recourse seeking individuals the algorithms can identify recourses. The only exception is admission data set for the NTK model, where the success rate lies at 60\%. Across all recourse methods we set $\lambda \to 0$. Note that the default implementations use early stopping once a feasible recourse has been identified. 

\begin{table}
\centering
\begin{tabular}{cc}
\toprule
Model  &  Parameters \\
\midrule
Linear Regression  &  OLS, no hyperparameters.\\
NTK Regression & $\beta=2$ (Admission), $\beta=5$ (other data sets) \\
Logistic Regression & L2-Regularization with $C=1.0$ \\
Kernel-LSSVM & Gaussian Kernel with $\gamma=1.0$  (see \cite{cawley2004fast}) \\
ANN & 2-Layer, 30 Hidden units, Sigmoid, $\alpha=10$ (L2-Regularization) \\
\bottomrule
\end{tabular}
\caption{Model hyperparameters used in this work \label{tab:hyperparams}}
\end{table}

\subsection{Details on the $\ell_0$ Regularizer}
Since an $\ell_0$ regularizer is computationally intractable for high-dimensional optimization problems, we have to resort to approximations.
One such approximation approach was recently suggested by \citet{yamada20a}.
The underlying idea consists of converting the combinatorial search problem to a continuous search problem over distribution parameters. 
To this end, recall our optimization problem from the main text:
\begin{align}
\bomega^* = \argmax_{\bomega \in \{0,1\}^{n}} ~ m(\bomega) - \eta \cdot \lVert \one - \bomega \rVert_0.
\label{eq:ell0_objective_appendix}
\end{align}
We will now introduce Bernoulli random variables $Z_i \in \{0,1\}$ with corresponding parameters $\pi_i$ to model the individual $\omega_i$. 
Instead of optimizing the objective above with respect to $\bomega$ we will optimize with respect to distribution parameters $\bm{\pi}$:
\begin{align}
\bm{\pi}^* = \argmax_{\bm{\pi} } ~ m(\mathbf{Z}(\bm{\pi})) -  \eta \cdot \lVert \one - \mathbf{Z}(\bm{\pi}) \rVert_0.
\label{eq:ell0_objective_surrogate_one}
\end{align}
Since the above optimization problem is known to suffer from high-variance solutions, \citep{yamada20a} suggest to use a Gaussian-based continuous relaxation of the Bernoulli variables:
\begin{align}
    \tilde{Z}_i = \max(0, \min(1, \mu_i + \epsilon_i)),
\end{align}
where $\epsilon_i = \mathcal{N}(0, \sigma^2)$, resulting in the following optimization problem:
\begin{align}
\bm{\mu}^* = \argmax_{\bm{\mu}} ~ m(\tilde{\mathbf{Z}}(\bm{\mu})) - \eta \cdot \lVert \one - \tilde{\mathbf{Z}}(\bm{\mu}) \rVert_0.
\label{eq:ell0_objective_surrogate_two}
\end{align}
At inference time, the optimal weights are then given by $\tilde{Z}_i^* = \max(0, \min(1, \mu_i^*)) ~ \forall i \in [n]$. 
To obtain discrete weights, we take the argmax over each individual $\tilde{Z}_i$.

\subsection{Details on the Jackknife Approximation}

\begin{figure}
\begin{algorithm}[H]
\label{alg:thresh}
\begin{algorithmic}
\caption{Greedy \recourseo invalidation}
\State \textbf{Required:} Model: $f_{\bw(\one)}$; Matrix of Recourses: $\Xc_{\one} \in \mathbb{R}^{q \times d}$; $d$: input dimension; $q$ number of recourse points on test set; $n$: \# train points; $M$: max \# deleted train points; $s$: invalidation target
\State $\bomega^{(0)}=\one_n$ \Comment{All training instances present}
\For{$m=1:M$} 
\State $\bomega^{(m)} \leftarrow \bomega^{(m-1)}$
\State $\tilde{\textbf{Y}} =  \mathbf{0}_{n \times q}$ \Comment{Recourse outcomes}
\State $\bm{J} = \mathbf{0}_{n \times q}$ \Comment{Invalidations present}
\State $S^{(m)} \leftarrow \left\{i ~\middle|~\bomega_i^{(m)} \neq 0 \right\}$ \Comment{Set of train instances present at iteration $m$} 
\For{$i \in S^{(m)}$} 
\State $\bw^{(i)}_{\text{new}} = \texttt{update\_w}(\bomega^{(m)}_{-i})$  
\Comment{$\bomega^{(m)}_{-i}$ has additionally set weight $i$ to 0.}
\State~\Comment{Use analytical or IJ solution for $\bw(\bomega)$}
\State $\tilde{\textbf{Y}}[i,:] =  f_{\bw^{(i)}_{\text{new}}}(\Xc_{\one})$ \Comment{New recourse outcomes}
\State $\bm{J}[i,:] = \mathbb{I}( \tilde{\textbf{Y}}[i,:] < s)$ \Comment{Invalidation present}
\EndFor
\State index $\leftarrow{}$ $\argmax_{i \in S^{(m)}} \lVert \mathbf{J}[i,:] \rVert_1$  \Comment{Find point that leads to highest invalidation}
\State $\bomega^{(m)}[\text{index}] = 0$ \Comment{Remove training point}
\EndFor
\State \textbf{return}: $\bomega^{(M)}$ \Comment{\dataweights indicating $M$ removals}
\label{algorithm:greedy_deletion}
\end{algorithmic}
\end{algorithm}
\end{figure}
When the \parameters $\bw$ are a function of the \dataweights by solving \eqref{eq:weighted_erm} we can approximate $\bw(\bomega)$ using the infinitesimal Jackknife (IJ) without having to optimize \eqref{eq:weighted_erm} repeatedly \citep{jaeckel1972infinitesimal,efron1982jackknife,giordano2019swiss,giordano2019highswiss}:
\begin{align}
\bw_{\text{IJ}}(\bomega) = \bw_{\one} - \bH_{\one}^{-1} \bG_{\bomega - \one},
\label{eq:jackknife_update}
\end{align}
where $\bG$ and $\bH_{\one}$ are the Jacobian and the Hessian matrices of the loss function with respect to the \dataweights evaluated at the optimal \parameters $\bw$, i.e., $\bG_{\bomega - \one} = \frac{1}{n} \sum_{i=1}^n  (\omega_i-1) \cdot \frac{\partial \ell(f_{\bw}(\bx_i), y_i)}{\partial \bw}$ and $\bH_{\one} = \frac{1}{n} \sum_{i=1}^n  \frac{\partial^2 \ell(f_{\bw}(\bx_i), y_i)}{\partial \bw \partial \bw^\top}$.
Note that this technique computes the Hessian matrix $\bH_{\one}$ only once. 
Using this Jackknife approximation, the Jacobian term $\bG_{\bomega - 1 }$ becomes an explicit function of the \dataweights which makes the Jackknife approximation amenable to optimization.

\begin{figure}
\begin{algorithm}[H]
\begin{algorithmic}
\label{alg:thresh}
\caption{SGD \recourseo invalidation}
\State \textbf{Required:} Model: $f_{\bw(\one)}$; Matrix of Recourses: $\Xc_{\one} \in \mathbb{R}^{q \times d}$; $d$: input dimension; $q$ number of recourse points on test set; $n$: \# train points; $M$: max \# deleted train points; $s$: invalidation target
\State $\bmu^{(1)}=\one_n$ \Comment{Mu are soft data weights that are opimized.}
\For{$m=1:\text{Step}$} \Comment{Perform $Step$ number of updates.}
\State $\delta{-}\text{loss}{=}0.0$
\For{$k=1:K$} \Comment Use $K$ Monte-Carlo Samples for the approximation
    \State Sample $\beps_k^{(m)} \sim \mathcal{N}\left(0,\sigma^2\bI_n\right)$
    \State $\bomega_k^{(m)} = \max\left(0, \min\left(1, \bmu^{(m)} +\beps_k^{(m)}\right)\right)$ \Comment{Sample (soft) data weights as in  \cite{yamada20a}}
    \State $\bw^{(m)}_{k,\text{new}} = \texttt{update\_w}(\bomega^{(m)}_k)$ \Comment{\parbox{5cm}{Compute model weights from data weights either analytically or with IJ}}
    \State $l^{(m)}_{k} = \text{sigmoid}\left(f_{\bw^{(m)}_{k, \text{new}}}(\Xc_{\one}) - s \right)$ \Comment{\parbox{5cm}{Predict with new weights and compute soft invalidation.}}
    \State $\delta{-}\text{loss} = \delta{-}\text{loss} +  \lVert l^{(m)}_{k}\rVert_1$ \Comment{Add up soft inval. loss}
\EndFor
\State $r^{(m)}  = \sum_{i=1}^n \Phi\left(\frac{1-(\bmu^{(m)})_i}{\sigma}\right)$ \Comment{Sparsity Regularizer from \cite{yamada20a}}
\State $\bmu^{(m+1)} =  \bmu^{(m)} + \gamma\nabla_{\bmu^{(m)}}\left(\frac{\delta{-}\text{loss}}{D}+ \lambda r^{(m)}\right)$ \Comment{Grad. Descent with lr. $\gamma$}
\EndFor

\State $\text{removed\_ind}$ = \texttt{argsort}$(\bmu^{(\text{Step}+1)})$ \Comment{Sort indices ascendingly}
\State $j=0$
\State $\bomega = \one_n$
\While{$j < M$} \Comment{Binarize and fulfil max number M} 
\State $\bomega$[removed\_ind[j]]= 0 
\State $j = j +1 $
\EndWhile
\State \textbf{return}: $\bomega$ \Comment{\dataweights indicating $M$ removals}
\label{algorithm:sgd_deletion}
\end{algorithmic}
\end{algorithm}
\end{figure}


\end{document}